\RequirePackage{fix-cm}

\documentclass[smallextended]{svjour3}       % onecolumn (second format)

\smartqed  % flush right qed marks, e.g. at end of proof

\usepackage{graphicx}
\usepackage{graphics}
\usepackage{color}
\usepackage{natbib}
\usepackage{amssymb}
\usepackage{mathtools}
\usepackage{algorithm, algorithmic}
\usepackage{enumerate}
\usepackage{booktabs}
\usepackage{multirow}
\usepackage[hidelinks]{hyperref}
\usepackage[caption=false]{subfig}
\usepackage{xcolor}
\usepackage{xcolor}

\usepackage{booktabs}
\allowdisplaybreaks

\newcommand{\dataset}[1]{{\fontfamily{cmtt}\selectfont #1}}
\begin{document}
	
	\title{Exploring dual information in distance metric learning for clustering  %\thanks{Grants or other notes
	}

	\author{Rodrigo Randel \and Daniel Aloise \and \\Alain Hertz	}

	\titlerunning{Exploring Dual Information in  Distance Metric Learning for Clustering }       % if too long for running head
	
	\institute{Rodrigo Randel  \at
		%D\'epartement de G\'enie Informatique et G\'enie Logiciel, 
		Polytechnique Montr\'eal, GERAD, Canada \\
		\email{rodrigo.randel@polymtl.ca}           %  \\
		\and Daniel Aloise \at
		%D\'epartement de G\'enie Informatique et G\'enie Logiciel, 
		Polytechnique Montr\'eal, GERAD,  Canada \\
		\email{daniel.aloise@polymtl.ca}      
		\and
		Alain Hertz \at
		%D\'epartement de Math\'ematiques et de G\'enie Industriel, 
		Polytechnique Montr\'eal, GERAD,  Canada \\
		\email{alain.hertz@polymtl.ca} 
		%\and
		%Simon J. Blanchard \at
		%McDonough School of Business, Georgetown University, USA \\
		%\email{sjb247@georgetown.edu}
		% \vspace{-1cm}    
	}

	\date{Received: date / Accepted: date}

	\maketitle

\vspace{-0cm}\begin{abstract}
	Distance metric learning algorithms aim to appropriately measure similarities and distances between data points. In the context of %{\blue semi-supervised}%
	clustering, metric learning is typically applied with the assist of side-information provided by experts, most commonly expressed in the form of cannot-link and must-link constraints. In this setting, distance metric learning algorithms move closer pairs of data points involved in must-link constraints, while pairs of points involved in cannot-link constraints are moved away from each other.  For these algorithms to be effective, it is important to use a distance metric that matches the expert knowledge, beliefs, and expectations, and the transformations made to stick to the side-information should preserve geometrical properties of the dataset. Also, it is interesting to filter the constraints provided by the experts to keep only the most useful and reject those that can harm the clustering process. To address these issues, we propose to exploit the dual information associated with the pairwise constraints of the semi-supervised clustering problem. Experiments clearly show that distance metric learning algorithms benefit from integrating this dual information.
\end{abstract}

\section{Introduction}

A large collection of data mining techniques strongly rely on the use of similarity or dissimilarity measures among data objects to successfully perform their associated tasks~\citep{Aggarwal2015}. Consequently, defining a metric capable of recovering the underlying data structure is a critical component for achieving good results in many pattern recognition problems~\citep{Chang2006}. The effects of an inappropriate dissimilarity metric are notably severe under the unsupervised learning paradigm, mainly due to the lack of background information to  evaluate classification performance~\citep{Xing2003}. This is the case, for example, for the \textit{clustering} problem, one of the most popular data mining tasks, which is aimed to identify hidden homogeneous subgroups, named clusters, from the data~\citep{Hansen1997}.

Whereas defining a good fitting dissimilarity metric is highly problem-specific~\citep{Xiang2008}, the user of a clustering model has often no evidence or external information to assess the quality of the dissimilarity metric adopted nor of the  clusters obtained. To mitigate this difficulty, \textit{distance metric learning} techniques emerged as a mechanism to automatically learn how to appropriately measure similarities and distances through the use of external knowledge about the data~\citep{bellet2015metric}. In clustering, distance learning frequently occurs under the semi-supervised paradigm, where domain experts are allowed to provide additional side-information regarding the data distribution.  Therefore, semi-supervised clustering methods aim to find solutions that are more in line with the expert knowledge, beliefs, and expectations.

 Typically, side-information is formulated by means of pairwise constraints, for which the user provides information regarding the relationship between a pair of data objects. In this sense, a \emph{must-link} constraint informs that two data points are similar and, therefore, must be assigned to the same cluster. Likewise, a \emph{cannot-link} constraint ensures that a pair of points must be assigned to different clusters. Pairwise constraints arise naturally in many applications, e.g., image retrieval~\citep{Aharon2005}, and, in many circumstances, are more practical to be obtained than  class-labels~\citep{Zhang2008}.

Accordingly, under the semi-supervised clustering paradigm with pairwise constraints, distance learning methods are designed to find transformations to the data features, so as to bring closer pairs of data objects involved in must-link constraints and to move away pairs of data objects involved in cannot-link constraints for the metric space under consideration.  Thus, distance metric learning can mitigate the effect of adopting a less appropriate clustering model to capture the underlying structure of the data. 

\begin{figure*}[!htb]
\centering

\subfloat[Ground-truth partition]{\includegraphics[width=0.48\linewidth, keepaspectratio,page=1]{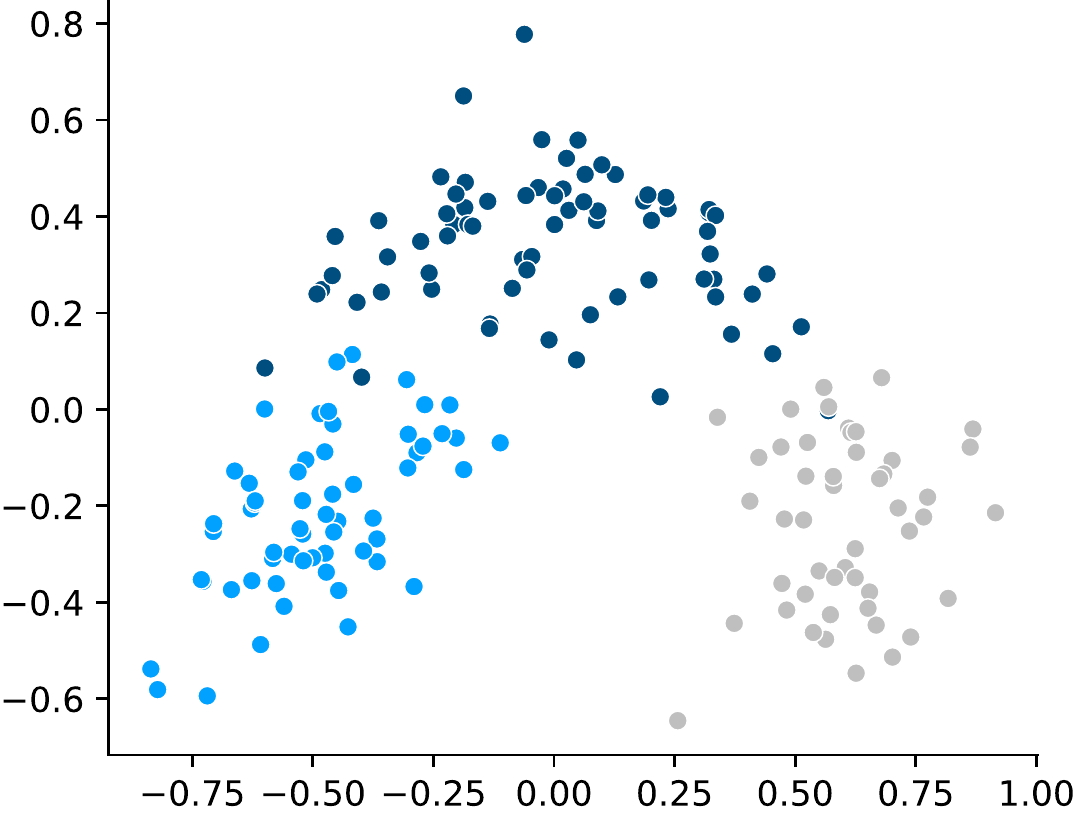}}\\
\subfloat[Output of MPCK-Means based on Euclidean distances]{\includegraphics[width=0.48\linewidth, keepaspectratio,page=2]{img/wine_wrong_distance.pdf}}
\hfill
\subfloat[Output of MPCK-means based on Cosine distances]{\includegraphics[width=.48\linewidth, keepaspectratio,page=3]{img/wine_wrong_distance.pdf}} 

\caption{Illustration of the use of MPCK-Means on the \dataset{Wine} dataset, with two metrics and 20 pairwise constraints.}
\label{fig:ex_wrong_metric}
\end{figure*}

Establishing a re-featured dataset through distance metric learning has its own challenges. 
For example, the choice of the metric to be used initially in order to define the notion of dissimilarity is crucial and has a significant impact on the resulting clustering solution. For illustration, an example is shown in Figure \ref{fig:ex_wrong_metric}, where clustering solutions  for the popular \dataset{Wine} dataset~\citep{Dua2019} are obtained with the distance metric learning algorithm  MPCK-Means~\citep{Bilenko2004}, using two different metrics and 20 randomly generated pairwise constraints. The data points are displayed with the two principal components.  We observe that the Euclidean distance is a more suitable metric for finding the ground-truth partition than the Cosine distance, even when the same set of pairwise constraints is used. This example illustrates the need to first find a suitable metric so that distance metric learning algorithms can actually benefit from the background knowledge.

Another issue with distance metric learning methods is that the transformations that are applied to the data points do not necessarily preserve the geometric properties of the dataset.
This can be problematic since these transformations are based on pairwise constraints collected from different sources than the ones that generated the dataset.
This occurs because domain experts provide pairwise constraints based on their beliefs and expectations, and these may not be consistent with the distribution of the data.
Such a defect is particularly apparent when the pairwise constraints fail to deal well with critical regions of the  space (e.g., overlapping of classes)~\citep{Lecapitaine2018}. Therefore, a method capable of keeping the modified space as close as possible to the original one is of great interest. This not only makes the interpretation of the clustering more reliable, 
but it also contributes to a more accurate view of the transformed dataset when represented using the learned metric.

It has been shown that some constraints can negatively affect the clustering quality~\citep{Davidson2006}.  Nevertheless, little is known about which constraints are best able to guide a semi-supervised clustering algorithm. Users of clustering algorithms are thus left with the sole option of using all the available constraints, which corresponds to a hit-or-miss game~\citep {Zhang2021}. A mechanism to assess the importance of pairwise constraints, to keep only those that have a positive effect, would therefore be of great benefit to users.

This research proposes to address the above issues by defining a methodology for measuring and analyzing the effects of pairwise constraints in a clustering model. To this end, we resort to a Lagrangian formulation of the clustering problem, where pairwise constraints are relaxed and transferred to the clustering objective function. This process introduces  Lagrange multipliers, called \emph{dual variables}, which indicate the price to be paid by the model in order to violate a constraint.
 In other words, by associating each pairwise constraint with a dual variable, we can measure the impact of each constraint in the clustering solution. This idea was introduced in our previous work~\citep{randel2020a} with the objective of identifying the presence of erroneous/contradictory pairwise constraints.
 Motivated by the wealth of information that can be acquired from dual variables, we propose in this paper to further exploit this knowledge by developing tools to help manage the problems described above  in  distance metric learning for clustering.

The rest of the paper is organized as follows. In Section \ref{sec:related_works}, we summarize prior research on   distance metric learning  for clustering, with supervision based on pairwise constraints. We then explain in Section \ref{sec:methodology} how to generate and exploit dual information obtained from pairwise constraints within clustering optimization problems.  The next three sections illustrate the use of the proposed tools: we show in Section \ref{sec:user_case1} how to identify the most appropriate metric in clustering models; in Section \ref{sec:user_case2}, we explain how to design distance metric learning methods so that the transformations that are applied to the data points preserve as much as possible the geometric properties of the dataset; we describe in Section \ref{sec:user_case3} a procedure that filters the most useful constraints and demonstrate how it can be integrated with a deep learning framework recently proposed in the literature. Concluding remarks are given in the last section. 

\section{Related works}
\label{sec:related_works}

In this paper, we focus on distance metric learning algorithms for clustering, with supervision based on pairwise constraints. The literature on metric learning is however much broader, and the reader is referred to \citet{bellet2015metric} for a thorough review on the subject.

 The distance metric learning task is often performed by assuming that dissimilarities are expressed as Mahalanobis distances~\citep{maha}.
 This distance has generated a lot of interest due to its flexibility and nice interpretation in terms of a linear projection~\citep{bellet2015metric,Nguyen2019}. Given a dataset $ O{=}\{o_1, \ldots, o_n\} $ of $ n $ points in the $d$-dimensional space, 
the objective of the Malahanobis distance  learning  is to determine a positive semi-definite $d{\times} d$ matrix $ \mathcal{A} $, such that the Mahalanobis distance between two points $ o_i $ and $ o_j $ is increased if a constraint imposes that they must belong to the same cluster, and is decreased if the points are to be in different clusters. The Mahalanobis distance between points $o_i$ and $o_j$ in $\mathbb{R}^d$  is defined as:
 \begin{align}
	d_{\mathcal{A}}(o_i, o_j) = \sqrt{(o_i- o_j)^{T}\mathcal{A}(o_i - o_j)}. \label{eq:mahalanobis}
\end{align}

The learned matrix $\mathcal{A} $ can be used to transform the dataset $O$ into a set $O'$ by defining $o^\prime_i = \mathcal{A}^{1/2}o_i$ for $i=1,\ldots,n$ (where $\mathcal{A}^{1/2}$ is the unique matrix $\mathcal{B}$ that is positive semidefinite and such that $\mathcal{B}\mathcal{B} = \mathcal{B}^T\mathcal{B} = \mathcal{A}$). The standard Euclidean distance can then be used on $O'$ by classical data mining algorithms. 
The seminal work exploring this idea was done by~\cite{Xing2003}. Formally, given a set $\mathcal{ML}$ of must-link constraints and a set $ \mathcal{CL}$ of cannot-link constraints, the \textit{Mahalanobis distance metric learning problem}  is expressed as follows, where $\mathbb{S}$ is the set of positive semidefinite $d{\times} d$ matrices :
\begin{align}
	&\begin{aligned}
		\max_{\mathcal{A} \in \mathbb{S}} &\sum_{(o_i,o_j)\in \mathcal{CL}} d_{\mathcal{A}}(o_i, o_j) \\
		\text{subject to} 
		&\sum_{(o_i,o_j) \in \mathcal{ML}} d_{\mathcal{A}}(o_i, o_j) \leq c	
	\end{aligned} \label{eq:model_mmc}	
\end{align} 

\noindent In words, the sum of the distances between dissimilar points is maximized, while for similar points, this sum must be less than a constant $c$ which is typically set equal to 1.
One could also choose to minimize the sum of the distances between similar points, while keeping the sum of distances between dissimilar points greater than $ c $.
Problem \eqref{eq:model_mmc} can be solved to optimality using a projected gradient descent algorithm. However, if $ \mathcal {A} $ is a full matrix, the time complexity of determining the eigenvalues of $ \mathcal {A} $ is $ O (d ^ 2) $, which is computationally too expensive for high-dimensional data~\citep{Bilenko2004, Wu2012}. To reduce this complexity, one can impose that $ \mathcal {A} $ be a diagonal matrix, which is equivalent to performing a feature weighting on the dataset.

A classical distance metric learning technique for clustering with pairwise constraints is the MPCK-Means algorithm~\citep{Bilenko2004}. Instead of computing a single matrix $ \mathcal{A} $, MPCK-Means determines one matrix per cluster. The method can thus define local transformations which allow  the clusters to have different shapes. This feature can be particularly useful to enforce specific shapes. For example, squared Euclidean distances used by the classical  \textit{k}-means heuristic are suitable for spheroidal clusters, but fail to represent dissimilarities when the clusters have other shapes.

We should also mention that the learning of nonlinear shapes can be achieved by using Bregman distances~\citep{Lei2009, Wu2012} and kernel functions~\citep{Baghshah2008, Kalintha2017, Nguyen2019}, both being more suitable than Mahalanobis distances for  high-dimensional data, or to work with datasets with nonlinear structures.

Methods based on deep learning have also been developed~\citep{hsu2015neural, Fogel2019, Zhang2021}. For example, \cite{Zhang2021} recently proposed to use the Deep Embedded Clustering (DEC) algorithm~\citep {xie2016unsupervised} followed by a training procedure that incorporates the violation of secondary constraints in the loss function.
The DEC method consists of first learning an embedded representation of the data points using an autoencoder network, and then executing a self-training routine to learn a clustering partition. To accomplish that, DEC uses the embedded points to define a \textit{soft} membership distribution (points can be partially assigned to more than one cluster), and then approximates it to a target distribution that resembles a \textit{hard} clustering membership (where data points are assigned to exactly one cluster) using the Kullback-Leibler (KL) divergence~\citep{kullback1951} loss function.  The work of \citet{Zhang2021}, extends the loss function by adding terms that accounts for diverse types of supervision, including pairwise constraints. 
It has been shown that the proposed algorithm outperforms previous semi-supervised clustering algorithms in terms of clustering accuracy.

\section{Dual information from the pairwise constraints}
\label{sec:methodology}

A $k$-partition of a set is a grouping of its elements into $k$ disjoint non-empty subsets called clusters. Given a set $O$ of $n$ data points and an integer $k$, we write $ \mathcal{P}(O,k)$  for the set of all $k$-partitions of $O$. Let $f: \mathcal{P}(O,k) \rightarrow \mathbb{R}$, be a function, usually called \emph{clustering criterion}, that assigns a value to every $k$-partition in $\mathcal{P}(O,k)$. The clustering problem is to determine a $k$-partition $P\in \mathcal{P}(O,k)$ with minimum (or maximum) value $f(P)$.

A widely used clustering criterion is the minimum sum of squared Euclidean distances from each data point to the centroid of its cluster, or minimum sum-of-squares clustering (MSSC) for short, which expresses both homogeneity and separation~\citep{Aloise2012}.  Let $ \mathcal{ML} $ and $ \mathcal{CL}$ be the  sets of pairs $(o_i,o_j)$ of data points involved in must-link and cannot-link constraints, respectively, where $(o_i,o_j)\in\mathcal{ML}\cup \mathcal{CL}$ only if $i<j$. The semi-supervised version of the clustering problem with MSSC can be expressed by means of the following optimization problem:
\begin{align}
	\mathop{\text{min}\quad}& \sum_{i=1}^{n}\sum_{c=1}^{k} x_{i}^{c}\|o_i - y_c\|^{2}\label{eq:clustering_obj}\\
	\text{subject to} \quad
	& \sum_{c=1}^{k}   x_{i}^{c}  = 1,  \quad  \forall i=1\dots n \label{eq:clustering_assignment}\\
	&   x_{i}^{c}  + x_{j}^{c}  \leq 1 \quad  \forall (o_i,o_j) \in \mathcal{CL},   \forall c = 1,\dots, k \label{eq:copcannotlink} \\
	&   x_{i}^{c}  - x_j^c = 0 \quad  \forall (o_i,o_j) \in \mathcal{ML}, \  \forall c = 1,\dots, k \label{eq:copmustlink} \\ 
	&   x_{i}^{c}  \in \{0,1\} \quad  \forall i=1,...,n,\  \forall c = 1,\dots, k.  \label{eq:clustering_end}
\end{align}

The feasible solutions to this problem correspond to  partitions $\{C_1,\ldots,C_k\}$ $\in \mathcal{P}(O,k)$
where every binary decision variable $x_{i}^{c} $ indicates whether data point $ o_i $ is assigned to cluster $ C_c $, and 
$ y_c $ is the centroid of cluster $ C_c $. Constraints \eqref{eq:clustering_assignment} ensure that each data point is assigned to exactly one cluster.  The cannot-link constraints are expressed by equations \eqref{eq:copcannotlink} and the must-link constraints by equations \eqref{eq:copmustlink}. To avoid situations where these constraints are satisfied with equality, we can replace them by the following equivalent constraints where $\epsilon$ is any real number in $]0,1[$:
\begin{align}
	&   x_{i}^{c}  + x_j^c \leq 1+\epsilon &\forall (o_i,o_j) \in \mathcal{CL}, \quad \forall c = 1,\dots, k \label{eq:copcannotlinkprime} \tag{\ref{eq:copcannotlink}$^\prime$}\\ 	&   x_{i}^{c}  - x_j^c \leq \epsilon &\forall (o_i,o_j) \in \mathcal{ML}, \quad \forall c = 1,\dots, k \label{eq:copmustlink1} \tag{\ref{eq:copmustlink}$^\prime$}\\ 
	&   x_{j}^{c}  - x_i^c \leq  \epsilon &\forall (o_i,o_j) \in \mathcal{ML}, \quad \forall c = 1,\dots, k. \label{eq:copmustlink2}  \tag{\ref{eq:copmustlink}$^{\prime\prime}$}
\end{align}

We aim to investigate how the pairwise constraints affect the clustering objective as expressed in \eqref{eq:clustering_obj}.
We believe that the cost of complying with a constraint may provide information on its importance. We thus attempt to answer the following questions:
\begin {itemize}
\item what are the most difficult constraints to satisfy?
\item is the adopted metric in agreement with the imposed constraints?
\item what are the most useful constraints for obtaining a low cost partition?
\end {itemize}

This constitutes a form of sensitivity analysis~\citep{PICHERY2014236} whose goal is to measure the impact of modifications of the input variables on the result of a clustering model. To obtain such information, we have recourse to the Lagrangian duality theory, as explained in \cite{randel2020a}

The Lagrangian function for the optimization problem \eqref{eq:clustering_obj}-\eqref{eq:clustering_end} can be obtained by introducing dual variables $ \eta_{ij}^{c}, \lambda_{ij}^{c} $ and $ \gamma_{ij}^{c} $ to penalize violations of inequality constraints \eqref{eq:copcannotlinkprime}, \eqref{eq:copmustlink1} and \eqref{eq:copmustlink2}. Specifically, the Lagrangian function $L(\eta, \lambda, \gamma)$ is defined as follows:

\vspace{-0.6cm}\begin{align}
		\quad L(\eta, \lambda, \gamma) = \min\quad &\sum_{i=1}^{n}\sum_{c=1}^{k} x_{i}^{c}\|o_i - y_c\|^{2}\nonumber\\ &+ \sum_{(o_i,o_j)\in \mathcal{CL}}\sum_{c=1}^{k}\eta_{ij}^{c}(1 + \epsilon -   x_{i}^{c}  - x_{j}^{c} )\nonumber\\
		&+ \sum_{(o_i,o_j)\in \mathcal{ML}}\sum_{c=1}^{k}\lambda_{ij}^{c}(\epsilon + x_{i}^{c}  - x_{j}^{c} )
		\nonumber\\&+ \sum_{(o_i,o_j)\in \mathcal{ML}}\sum_{c=1}^{k}\gamma_{ij}^c(\epsilon + x_{j}^{c}  - x_{i}^{c} ) \label{eq:lagrangian_start} \\
	 \text{subject to } \quad
	&  { \sum_{c=1}^{k}   x_{i}^{c}  = 1, \quad \forall i=1\dots n} \\
	& { x_{i}^{c}  \in \{0,1\} \quad \forall i=1,...,n;\quad \forall c = 1,\dots, k \label{eq:lagrangian_end}}
\end{align}

{\noindent and the dual of the integer program  \eqref{eq:clustering_obj}-\eqref{eq:clustering_end} can be expressed as follows:
\begin{align}
	L_{D}= \max_{ \eta,\lambda,\gamma\leq 0} L(\eta, \lambda, \gamma).
	\label{eq:dual_lagrange}
\end{align}}

The dual variables $\eta_{ij}^{c}, \lambda_{ij}^{c} $ and $ \gamma_{ij}^{c} $ provide an estimation of the cost paid by the clustering optimization model for respecting the associated constraints. 
If a pairwise constraint is inherently satisfied by \eqref{eq:clustering_obj}-\eqref{eq:clustering_end}, its associated dual variable should be zero. On the other hand, if a constraint is necessary to obtain an optimal solution to \eqref{eq:clustering_obj}-\eqref{eq:clustering_end}, a penalty must be paid for its violation, which implies that the associated dual variable should be strictly negative. 

A sub-gradient optimization algorithm can be used to solve \eqref{eq:dual_lagrange} and thus determine optimal values of the dual variables. The weak duality theorem~ \citep{Bertsimas1997,Fisher2004} asserts that $L_{D}$ is the best possible lower bound on the optimal value of \eqref{eq:clustering_obj}-\eqref{eq:clustering_end}. Consequently, the optimal values of the dual variables are reliable estimates of the price to be paid for complying with the constraints associated with them.
For more details,  the reader is referred to \cite{randel2020a}.

Formulation \eqref{eq:clustering_obj}-\eqref{eq:clustering_end} is based on the MSSC clustering criterion.  Neverthe- less, the proposed methodology is not limited to this particular objective. In fact, the strategy is flexible to be used in conjunction with several other clustering criteria and models represented in mathematical programming language~\citep{Hansen1997}.
To illustrate that, we use the k-medoids model~\citep{kaufman2009finding}. More precisely, let $D$ be a matrix, where each entry $ d_{ij} $ indicates the dissimilarity between data points $ o_i $ and $ o_j $. The medoid of a cluster is defined as the data point in the cluster whose average dissimilarity to all the data points in the cluster is minimal. The $k$-medoids problem is to determine a partition in $\mathcal{P}(O,k)$ that minimizes the sum of the distances of the data points to the medoid of their cluster.
The semi-supervised version of this clustering problem can be expressed by means of the following optimization problem:
\begin{align}
\min \quad &\sum_{i=1}^{n} \sum_{c=1}^{n}x_{i}^{c}d_{ic}
\label{eq:model_Medoids_start}\\
\text{subject to} \quad  
& \sum_{c=1}^{n} x_{i}^{c} = 1, &&\forall i = 1,...,n \label{eq:const_assigments} \\
& x_{i}^{c} \leq y_{c} &&\forall i=1,\ldots,n, \forall c=1,...,n \label{eq:const_openMedoids}\\
& \sum_{c=1}^n y_c = k \label{eq:const_kMedoids}\\
&   x_{i}^{c}  + x_{j}^{c}  \leq 1 &&\forall (o_i,o_j) \in \mathcal{CL}, \forall c = 1,\dots, n \label{eq:kmd_copcannotlink} \\
&   x_{i}^{c}  - x_j^c = 0 &&  \forall (o_i,o_j) \in \mathcal{ML}, \forall c = 1,\dots, n \label{eq:kmd_copmustlink} \\ 
& x_{i}^{c} \in \{0,1\} &&\forall i=1,\ldots,n, \forall c =1,\ldots,n, \label{eq:const_binaryVar} \\
& y_c \in \{0,1\} &&\forall c=1,\ldots,n, \label{eq:model_Medoids_end}
\end{align}

\noindent where every binary variable $ y_c $  indicates whether the data point $ o_c $ is one of the $k$ medoids, and  every binary variable $x_{i}^{c}$ indicates whether data point $o_i$ is assigned to a cluster having $o_c$ as medoid.
The Lagrangian function $L(\eta, \lambda, \gamma)$ associated with this optimization problem is then defined as in the previous case by replacing constraints \eqref{eq:kmd_copcannotlink} and \eqref{eq:kmd_copmustlink} by penalty terms in the objective function and the same sub-gradient optimization technique as in \citep{randel2020a} can be used to determine
$L_{D}= \max_{ \eta,\lambda,\gamma\leq 0} L(\eta, \lambda, \gamma)$.

\section{Identifying an appropriate dissimilarity measure}
\label{sec:user_case1}

Although the choice of the most appropriate dissimilarity metric is problem-specific,
users of an algorithm or a clustering model often have very little information to support their choice which can have a significant impact on the analysis of the quality of a solution. For example,
the homogeneity of a cluster depends on the similarity measure between the objects which is defined using the chosen metric.

 Distance learning methods are also highly dependent on the chosen distance metric.
 For example, the most commonly used dissimilarity measure for quantitative data is the Euclidean distance~\citep{Wu2012}. It assumes that each data feature is independent and equally important, which is not necessarily the case for the considered dataset. An alternative is to weight the relative importance of the data features. For instance, in the weighted Euclidean Distance-Based approach~\citep{Rao2013}, data features are  weighted according to their discriminating effect. These weights are generally set based on user preferences and interpretation, which is a subjective task prone to error.
 One can also choose to use secondary information such as pairwise constraints to improve the initial dissimilarities calculated from Euclidean distances, but this might not compensate for the fact that a more suitable metric exists for the underlying clustering optimization model. 

To tackle this issue, 
we propose to use a \textit{fitting score} which measures the adequacy of a metric with a given clustering model. This score is therefore a tool that helps experts to identify the metric that most closely matches their knowledge and beliefs. 
The fitness score $F(m)$ of a metric $ m $ is defined as
\begin{align}
F(m) = \displaystyle\sum_{(o_i,o_j)\in \mathcal{CL}}\;\sum_{c=1}^k \delta(\eta_{ij}^{c} =0)+\sum_{(o_i,o_j)\in \mathcal{ML}}\;\sum_{c=1}^k \delta(\lambda_{ij}^{c} =\gamma_{ij}^c = 0),  \label{eq:score}
\end{align}
where $\eta_{ij}^{c}, \lambda_{ij}^{c} $ and $ \gamma_{ij}^{c} $ are the optimal values of the dual variables resulting from the optimization of $L_D$ in \eqref{eq:dual_lagrange}, and 
$ \delta(e)  = 1 $ if expression $ e $ is true. 
The metric with the highest fitness score should be preferred for the given clustering model, as more pairwise constraints are inherently respected.

In the following subsections, we validate the ability of the fitness score to detect the most adequate metric for a clustering problem. Experiments are conducted on synthetic and real-world datasets.

\subsection{Validation of the fitness score for synthetic datasets}\label{sec:validationsynthetic}

In order to assess the usefulness of the proposed fitness score, we selected four of the most commonly used distance metrics in clustering problems, namely the Euclidean, Manhattan, Chebyshev, and Mahalanobis distances.

For each of the four metrics, we generated a two-dimensional dataset with 200 points and three balanced clusters $C_1,C_2,C_3$. This is how we proceeded.
\begin{itemize}
    \item For the Euclidean, Manhattan and Chebyshev distances, 
    we first generated 1000 pairs of points with coordinates drawn from a normal distribution $\mathcal{N}(10,1)$, and determined the maximum distance $\Delta $ (using the chosen metric) between the two points of a pair. 
    We then repeatedly generated three points $c_1,c_2,c_3$ with coordinates drawn from a normal distribution $\mathcal{N}(10,1)$, until $\min\{d(c_1,c_2),d(c_1,c_3),d(c_2,c_3)\}\geq \Delta/3$, where $d$ is the considered distance metric. The three resulting points were defined as the centers of the three clusters. Starting from $C_1{=}C_2{=}C_3{=}\emptyset$, we then iteratively generated points $p$ with coordinates drawn from a normal distribution $\mathcal{N}(10,1)$, determined the index $i$ such that $d(p,c_i)=\min\{d(p,c_1),d(p,c_2),d(p,c_3)\}$, and added $p$ to $C_i$ if $|C_i|{<}67$. The process stopped when $|C_1|{+}|C_2|{+}|C_3|{=}200$.
    \item For the Mahalanobis distance, we have defined $c_1{=}(50,0)$,  $c_2{=}(50,5)$ and $c_3{=}(50,10)$ as centers of the three clusters. 
    The 200 data points were then generated so that 
    every $x$ coordinate is a random number chosen  from the uniform distribution on the interval [0,100], and the $y$ coordinate of every data point in $C_i$ ($i=1,2,3$) is a random number generated from the uniform distribution on the interval $[5(i{-}1){-}0.1,5(i{-}1){+}0.1]$.
\end{itemize}

\begin{figure}[!htb]
	\centering
	\subfloat[Euclidean dataset]{\includegraphics[width=0.49\linewidth,height=3.8cm]{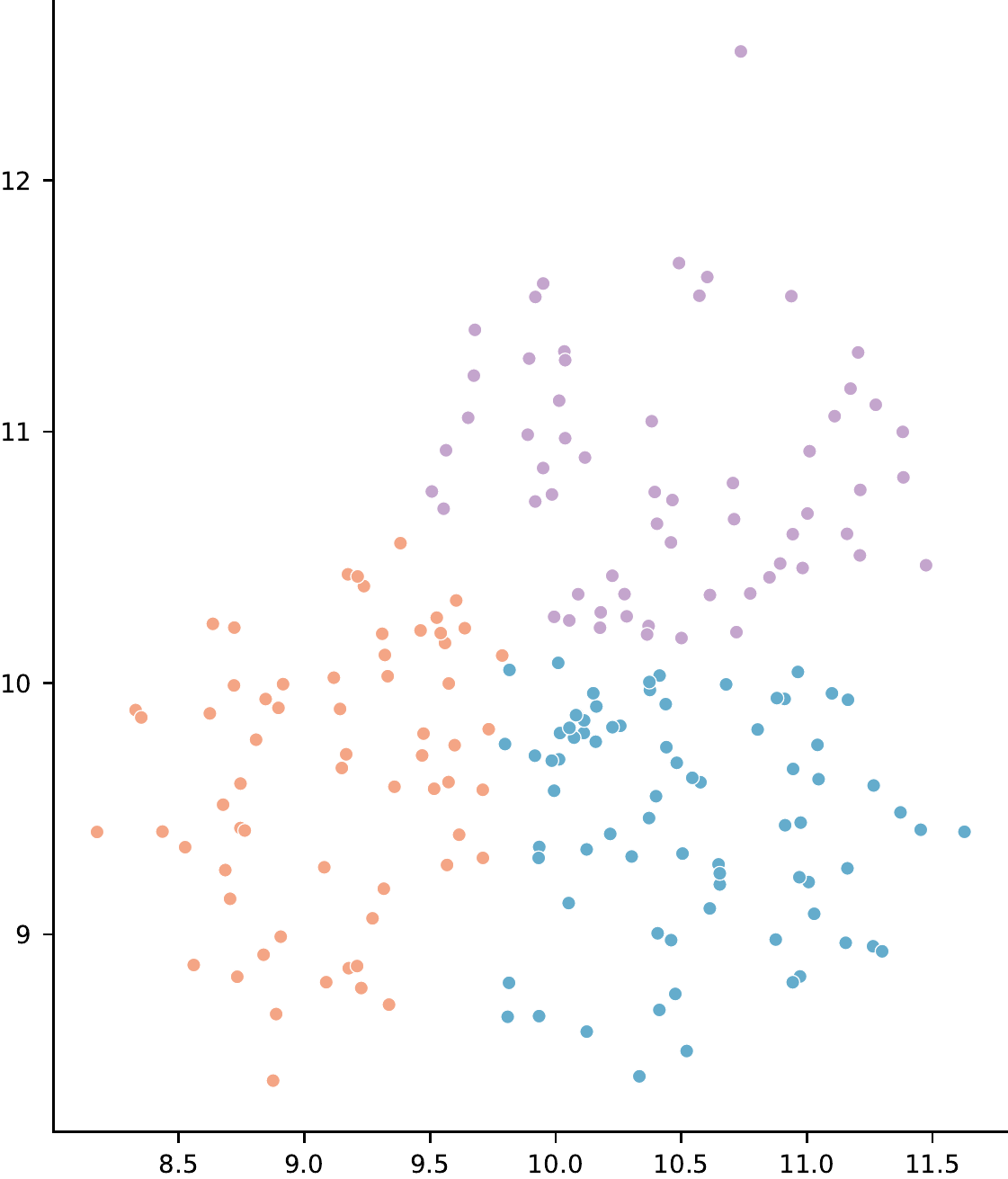}}\hfil
	\subfloat[Manhattan dataset]{\includegraphics[width=0.49\linewidth, height=3.8cm]{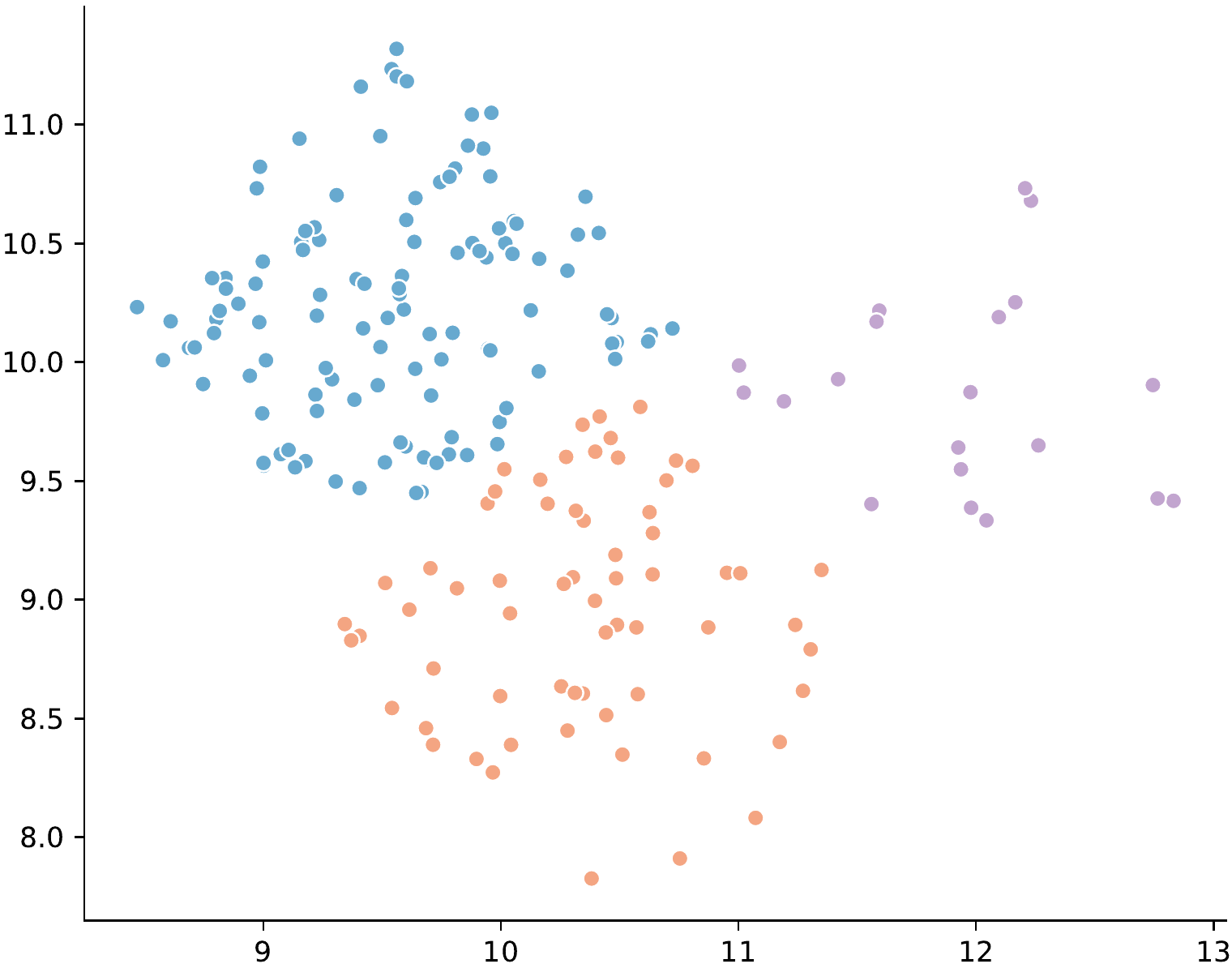}}\\
	\subfloat[Chebyshev dataset]{\includegraphics[width=0.49\linewidth, height=3.8cm]{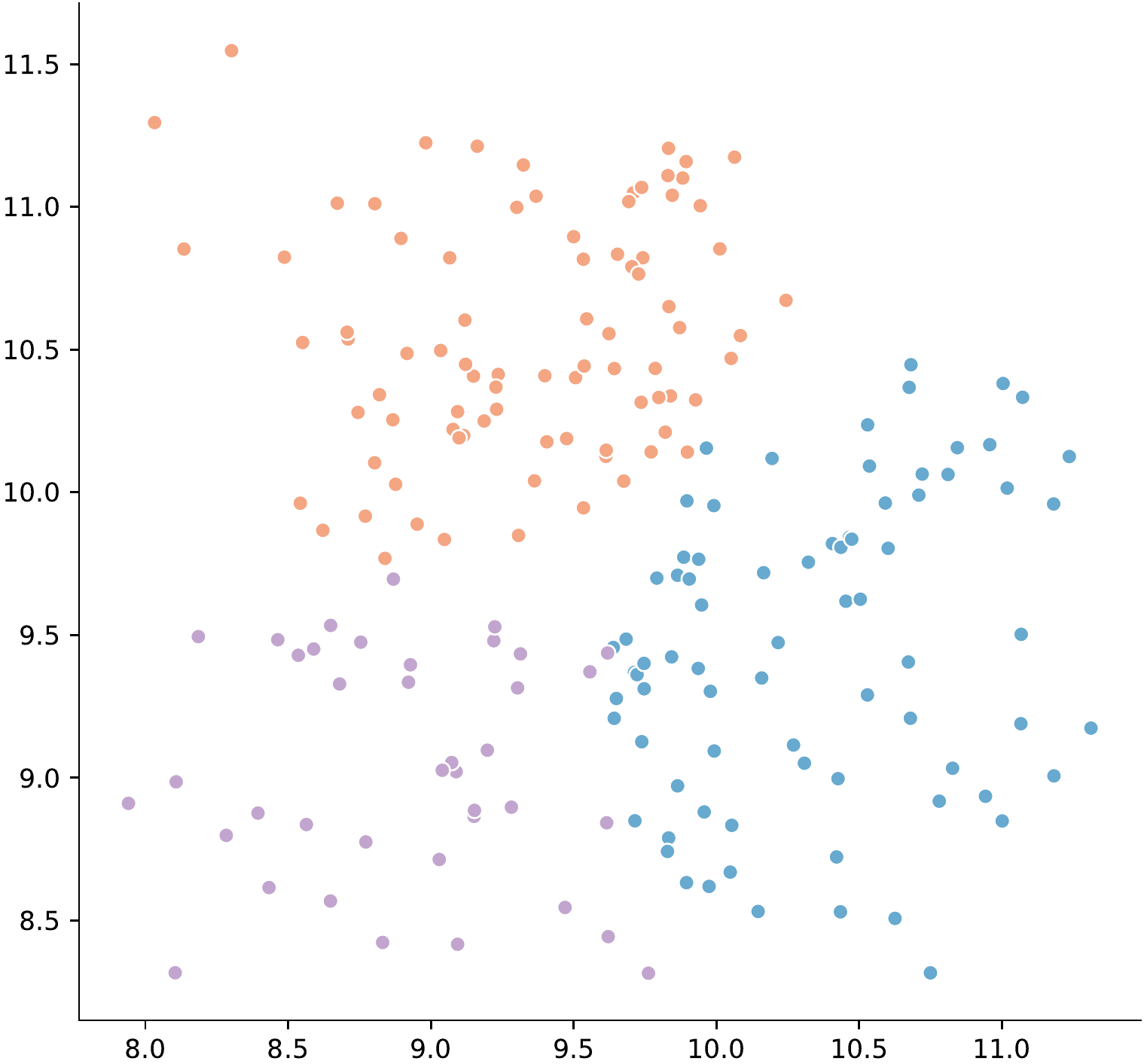}}\hfil
	\subfloat[Mahalanobis dataset]{\includegraphics[width=0.49\linewidth, height=3.8cm]{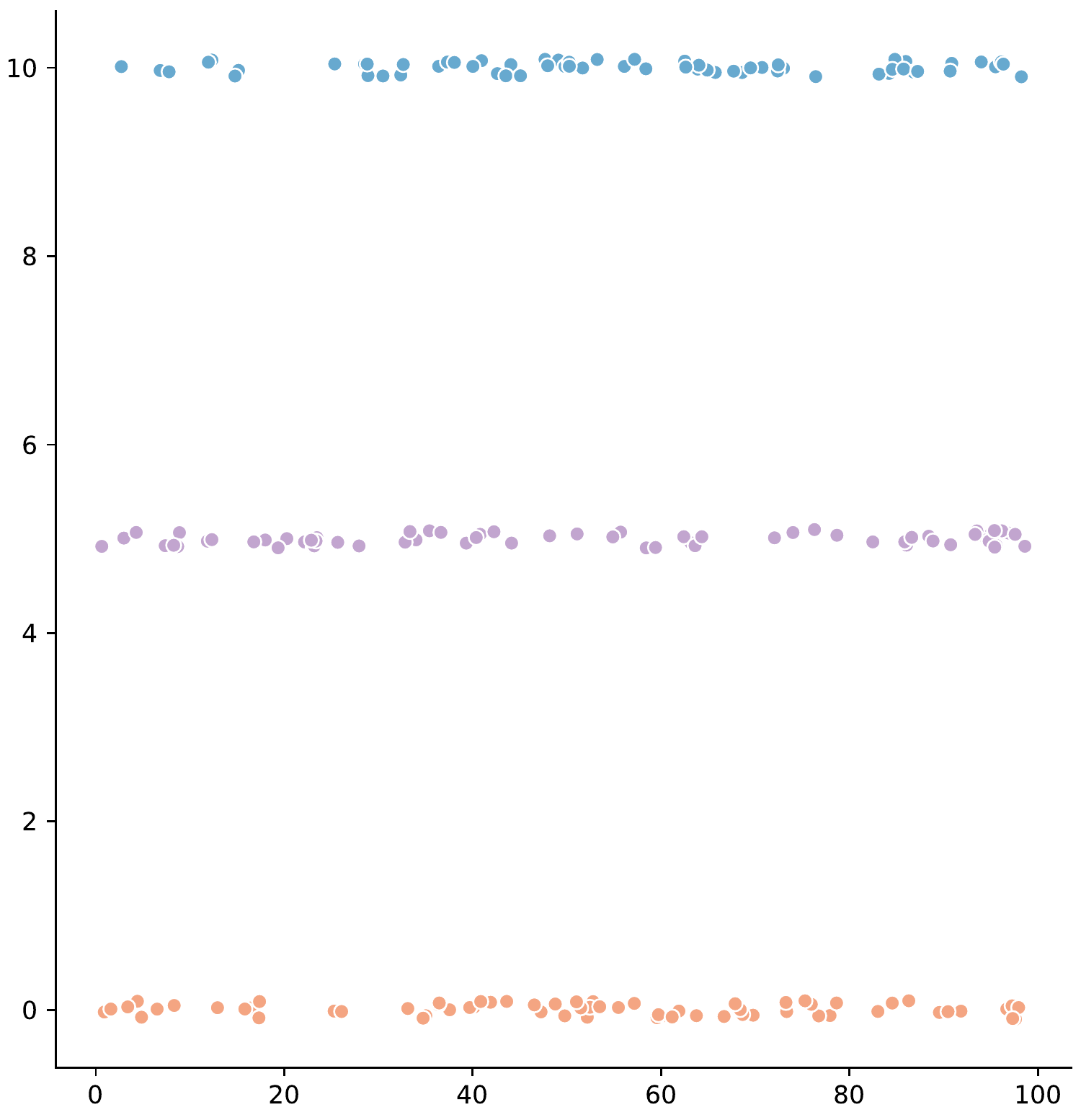}}
	
	\caption{Four synthetic datasets for four distance metrics, with their ground-truth partitions.}
	\label{fig:synthetic_data} 
\end{figure}

The four generated datasets and their ground-truth partitions are shown on Figure \ref{fig:synthetic_data}. Pairwise constraints were obtained using the \textit {boosting aggregating framework}~\citep {Breiman1996}.
For each of the 4 datasets, we generated 500 sets $E_1,\ldots,E_{500}$ of constraints, using the ground-truth partition. Each set $E_i$ contains $n_i$ constraints, with $n_i$ chosen randomly according to a uniform distribution in $\{1,\ldots, 100\}$.

We take advantage of the flexibility of the $k$-medoids model to solve the unsupervised clustering problem associated with each of the four datasets and each of the four metrics. More precisely, we set $d_{ij}$ equal to the distance between $o_i$ and $o_j$, using the considered metrics, and use CPLEX 12.8 to solve model
\eqref{eq:model_Medoids_start}-\eqref{eq:model_Medoids_end} without constraints \eqref{eq:kmd_copcannotlink} and \eqref{eq:kmd_copmustlink}.
The clustering accuracy of the resulting solutions is obtained using the Adjusted Random Index (ARI)~\citep{Hubert1985}, which measures the similarity between the obtained clustering solutions and the ground-truth partition.

\begin{figure}[!htb]
	\centering
	\subfloat[Euclidean dataset]{\includegraphics[width=0.50\linewidth, keepaspectratio,page=1]{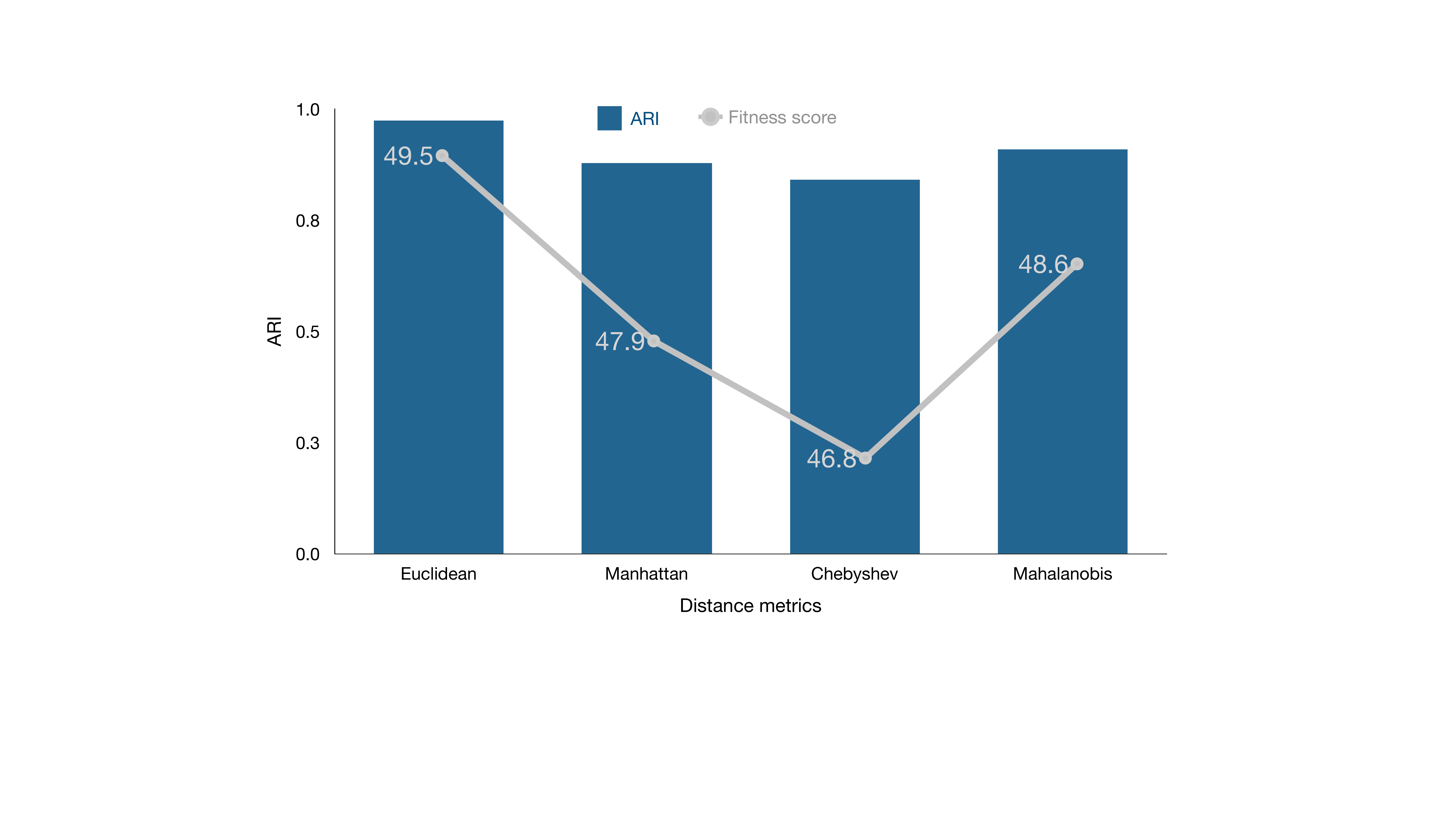}}\hfil
	\subfloat[Manhattan dataset]{\includegraphics[width=0.50\linewidth, keepaspectratio,page=2]{img/results_synthetic.pdf}}\\
    \subfloat[Chebyshev dataset]{\includegraphics[width=0.50\linewidth, keepaspectratio,page=3]{img/results_synthetic.pdf}}\hfil
    \subfloat[Mahalanobis dataset]{\includegraphics[width=0.50\linewidth, keepaspectratio,page=4]{img/results_synthetic.pdf}}
    
	\caption{Fitness scores and ARI for  synthetic datasets.}
 	\label{fig:results_synthetic}
\end{figure}

For each dataset, each set $E_i$ of pairwise constraints, and each metric $m$, we ran the sub-gradient algorithm of \cite{randel2020a} to solve \eqref{eq:dual_lagrange} and so produce optimal dual variables which give a fitness score as defined in \eqref{eq:score}. A time limit of 10 seconds has been allocated to each execution of the sub-gradient algorithm.  We report in Figure \ref{fig:results_synthetic}, for each dataset and each of the four metrics, the average fitness score calculated over the 500 sets of constraints. We also indicate the ARI values of the solutions produced by the unsupervised clustering algorithm based on the $k$-medoids model.

We observe that the fitness score is able to identify the most suitable metric for all datasets. 
The fluctuation of the fitness score agrees with that of the ARI and the best score is always obtained with the appropriate metric. Although the experiments were carried out on synthetic data, the results demonstrate that the optimal dual values of \eqref{eq:dual_lagrange} are useful for exploiting the information provided by the pairwise constraints, in order to suggest the metric which seems to adhere the most to the data.

\subsection{Validation of the fitness score for five real-world datasets}

We now consider five real-world datasets, the first four being available on the UCI repository~\citep{Dua2019}, and the fifth being described in~\citet{muller2020a}. Their characteristics are summarized in Table \ref{tab:real_data}. 

\begin{table}[!htb]
	\centering
	\caption{Real-data Applications for Evaluating the Score.}
	\begin{tabular}{l *3r l}
		Dataset& Samples & Classes & Features & Type \\
		\toprule
		\dataset{Iris} & 150 & 3 & 4 & Quantitative  \\
		\dataset{Wine} & 178 & 3 & 13 & Quantitative \\
		\dataset{Control} & 600 & 6 & 60 & Quantitative; time series \\
		\dataset{Twenty newsgroup} & 600  & 4 &- & Text \\
		\dataset{Eclipse bug report} & 460 & 5 & - & Discrete sequence; text\\
		\bottomrule
	\end{tabular}
	\label{tab:real_data}
\end{table}

\subsubsection{Quantitative datasets}

For the  quantitative datasets \dataset{Iris} and \dataset{Wine}, we used the same distance metrics as those carried out for the synthetic datasets of Section \ref{sec:validationsynthetic}, that is, Euclidean, Manhattan, Chebyshev, and Mahalanobis. 

For the \dataset{Control} dataset, each data point represents a time series composed of 60 values. Each time series was decomposed into six segments of 10 consecutive values, and we have set $o_i=(o_i^1,\ldots,o_i^6)$, where $o_i^k$ $(k=1,\ldots,6$) are the segments of $o_i$. The distance between two data points $o_i$ and $o_j$ was then defined as $\sum_{k=1}^{6}d(o_i^k,o_j^k)$, where $d$ is the considered metric.

\begin{figure}[!htb]
	\centering
	\subfloat[\dataset{Iris} dataset]{\includegraphics[width=0.4\linewidth, keepaspectratio,page=1]{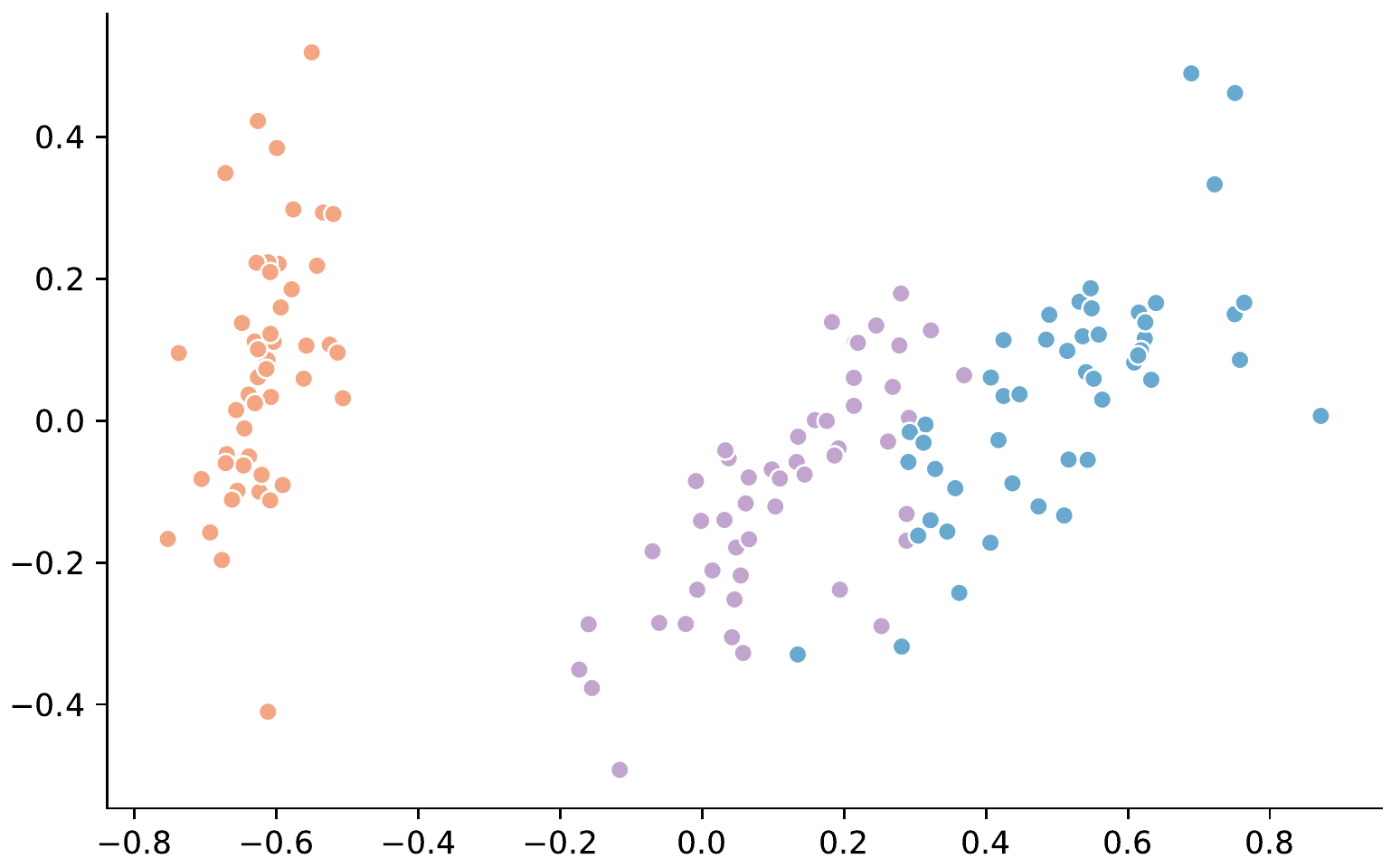}} \hfil
	\subfloat[Results for \dataset{Iris}]{\includegraphics[width=0.6\linewidth, keepaspectratio,page=1]{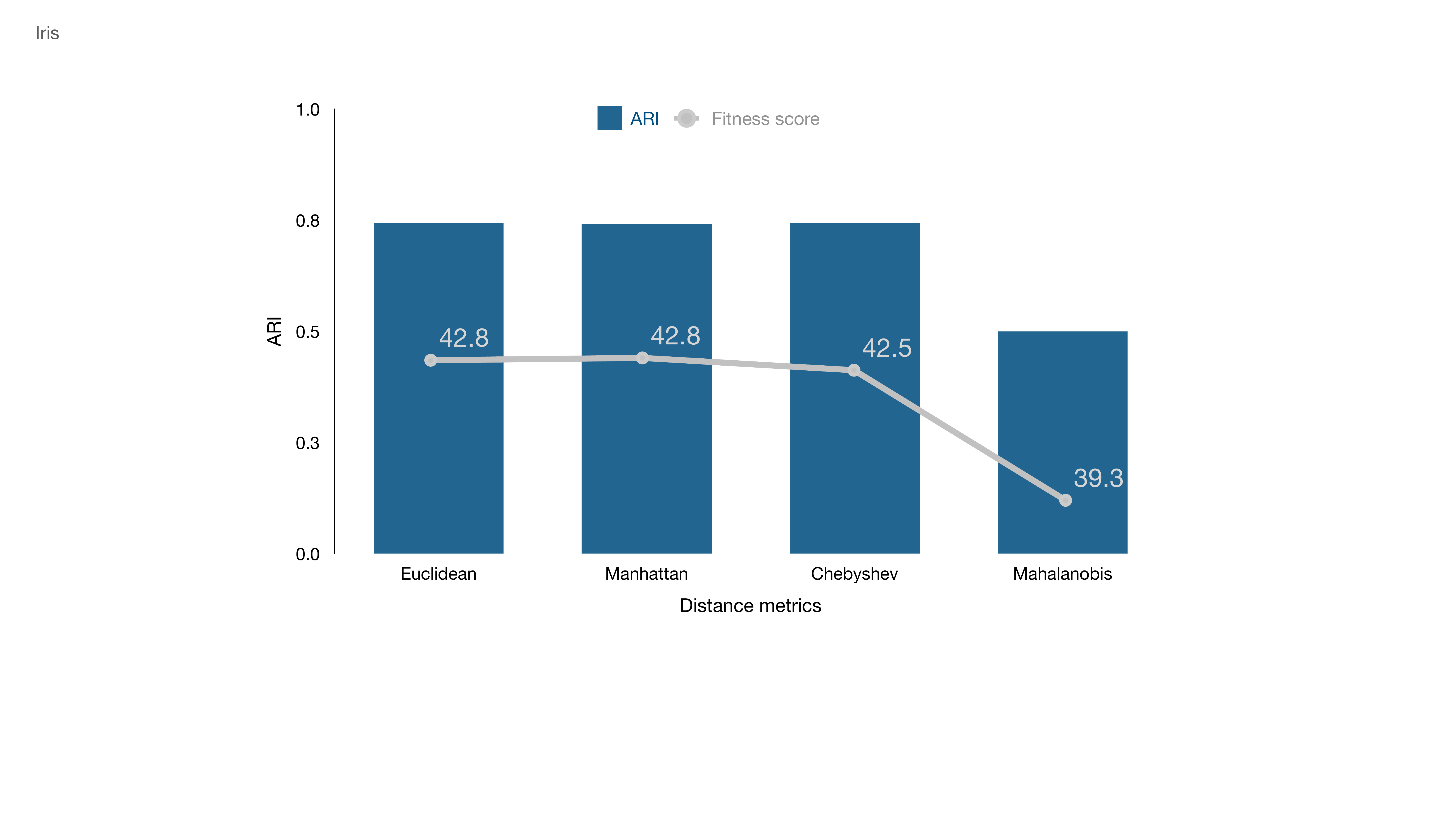}}\\
	\subfloat[\dataset{Wine} dataset]{\includegraphics[width=0.4\linewidth, keepaspectratio,page=1]{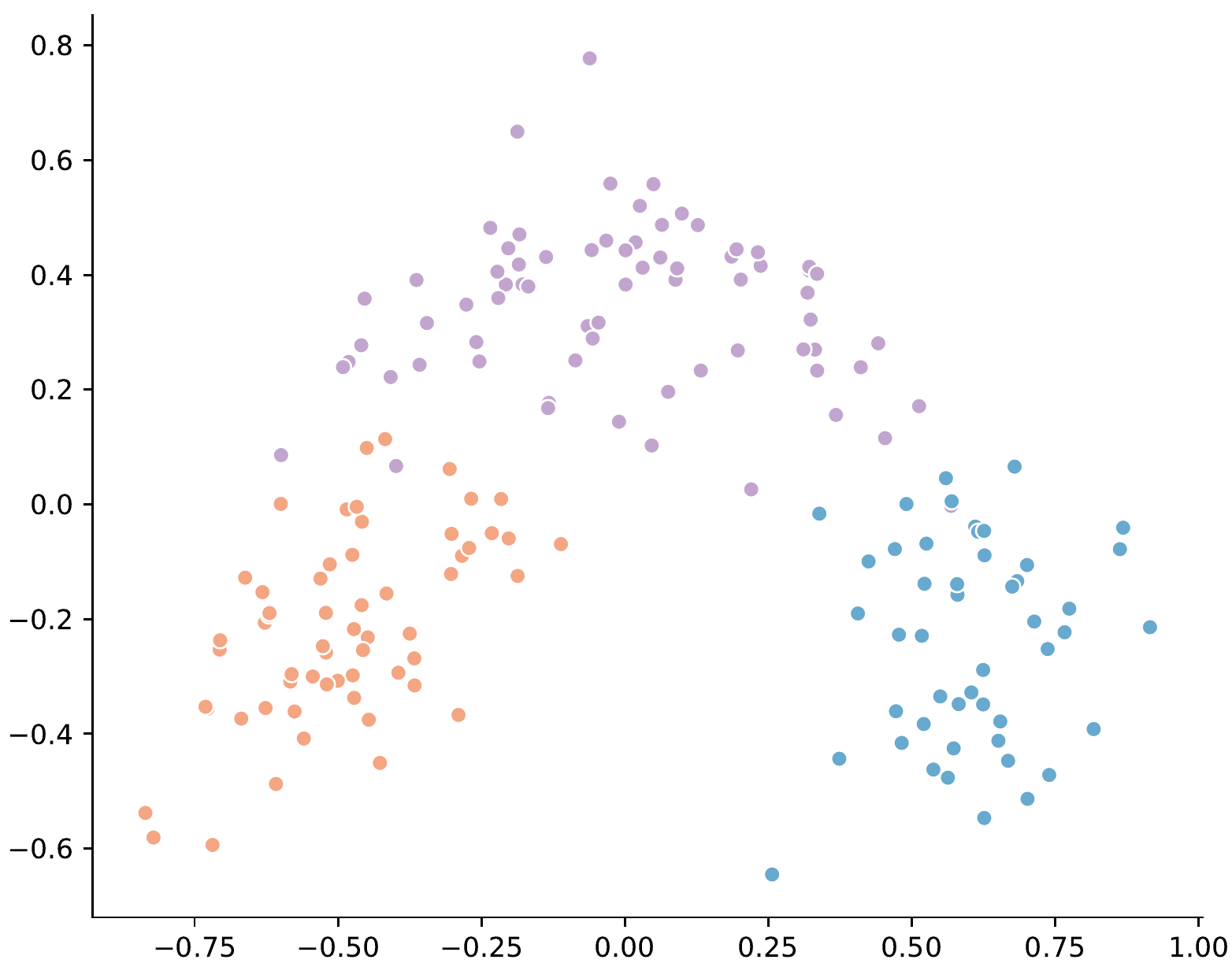}}\hfil
	\subfloat[Results for \dataset{Wine}]{\includegraphics[width=0.6\linewidth, keepaspectratio,page=2]{img/results_real.pdf}} \\
	\subfloat[\dataset{Control} dataset]{\includegraphics[width=0.4\linewidth, height=3cm,page=1]{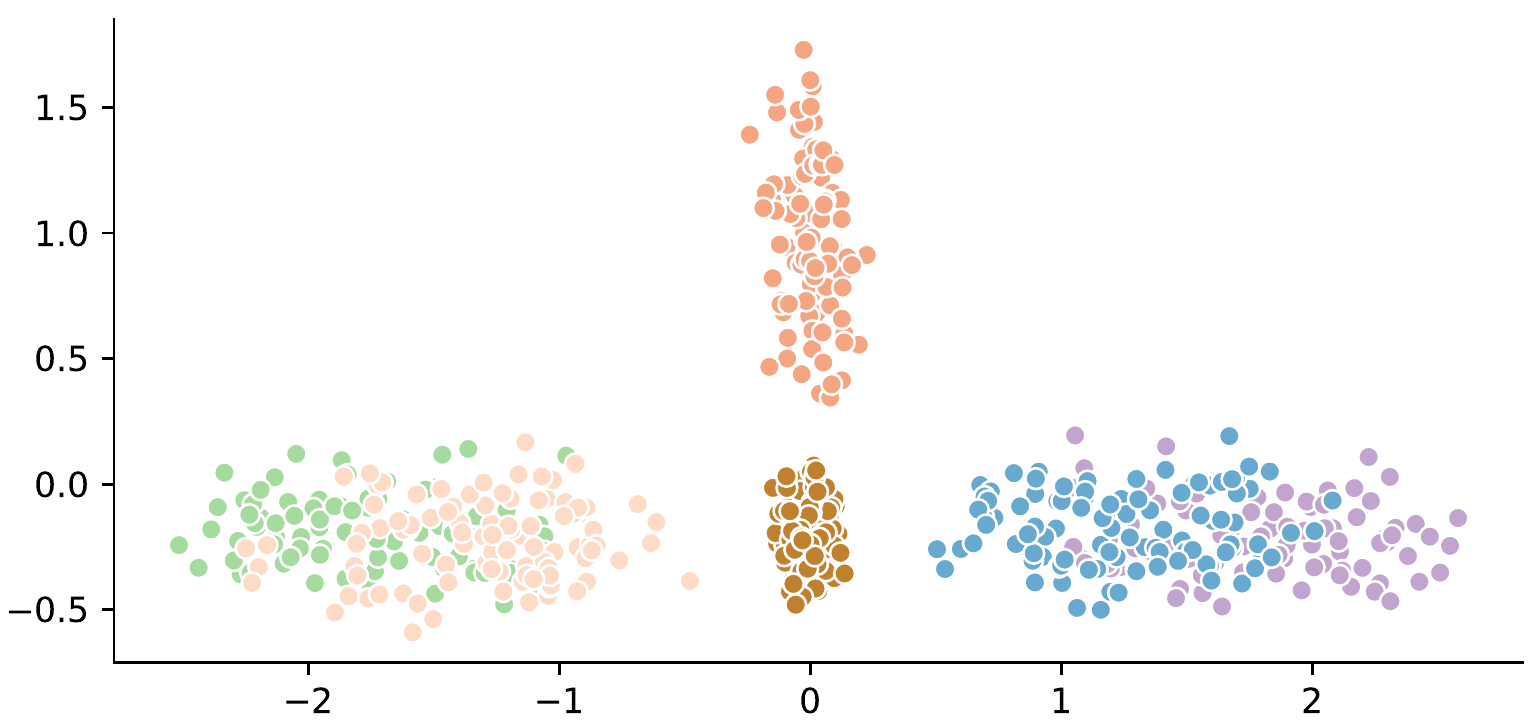}}\hfil
	\subfloat[Results for \dataset{Control}]{\includegraphics[width=0.6\linewidth, keepaspectratio,page=5]{img/results_real.pdf}}
	\caption{Ground-truth partition, fitness score and ARI for three quantitative datasets.}
	\label{fig:results_quantitative}
\end{figure}

For each dataset, we generated 500 constraints sets from the available ground-truth partitions, using the \textit {boosting aggregating framework}~\citep {Breiman1996}, as was done for the synthetic dataset.
The time limit for the sub-gradient algorithm has been set at 10 seconds for the \dataset{Iris} and \dataset{Wine} instances, and at 30 seconds for the \dataset{Control} dataset. 
The results are shown in Figure \ref{fig:results_quantitative}, alongside the representation of the datasets with the two principal components and their ground-truth partitions.

We observe that the $L_p$ norm-based distances (i.e., the Euclidean, Manhattan and Chebyshev distances) 
give similar results, both in terms of clustering accuracy and fitness score. This is not surprising given the popularity of these types of metrics for quantitative data clustering.

\subsubsection{Textual data}

The datasets \dataset{Twenty newsgroup} and \dataset{Eclipse bug report}  are composed of text samples. As typically done in document classification tasks, the texts were first converted to quantitative data through the use of a bag of words. For both datasets, we defined 10,000 features, which are originated from the most frequent terms in the dataset vocabulary. The direct use of $L_p$ norm-based distances on raw textual data is not recommended due to data sparsity and high-dimensionality. Besides, they are prone to be inconsistent if not normalized because 
 the distance between two long documents is very often larger than that between two short documents even if the two long documents are similar, and the short documents are unrelated~\citep{Aggarwal2015}. 
In contrast, the \emph{cosine distance} works by computing the angle between data samples, and is a popular choice for varied length-sized document classification.

In addition, instead of working with the raw frequency of the terms, we enhanced the features by calculating the \textit{term frequency–inverse document frequency} (TF-IDF)~\citep[e.g.][]{manning1999foundations}  to use global statistical measures to improve the dissimilarity computation. TF-IDF is based on the principle that documents matching with respect to rare terms are more likely to be similar than documents sharing many common terms (e.g. \emph{the}, \emph{like}).
In our experiments, we considered every data point as a vector in $\mathbb{R}^{10,000}$, where each coordinate indicates the TF-IDF of the corresponding term.

The \emph{Damerau–Levenshtein} distance is commonly used to establish the dissimilarity between discrete sequences. It measures the minimum number of operations necessary to change one sequence into another. Since the \dataset{Eclipse bug report} dataset is structured as a discrete sequence of stack trace functions, we decided to include this other distance in our tests.
A time limit of 30 seconds has been allocated to each execution of the sub-gradient algorithm.

\begin{figure}[]
	\centering
	\subfloat[\dataset{Twenty newsgroup} dataset]{\includegraphics[width=0.4\linewidth, keepaspectratio,page=1]{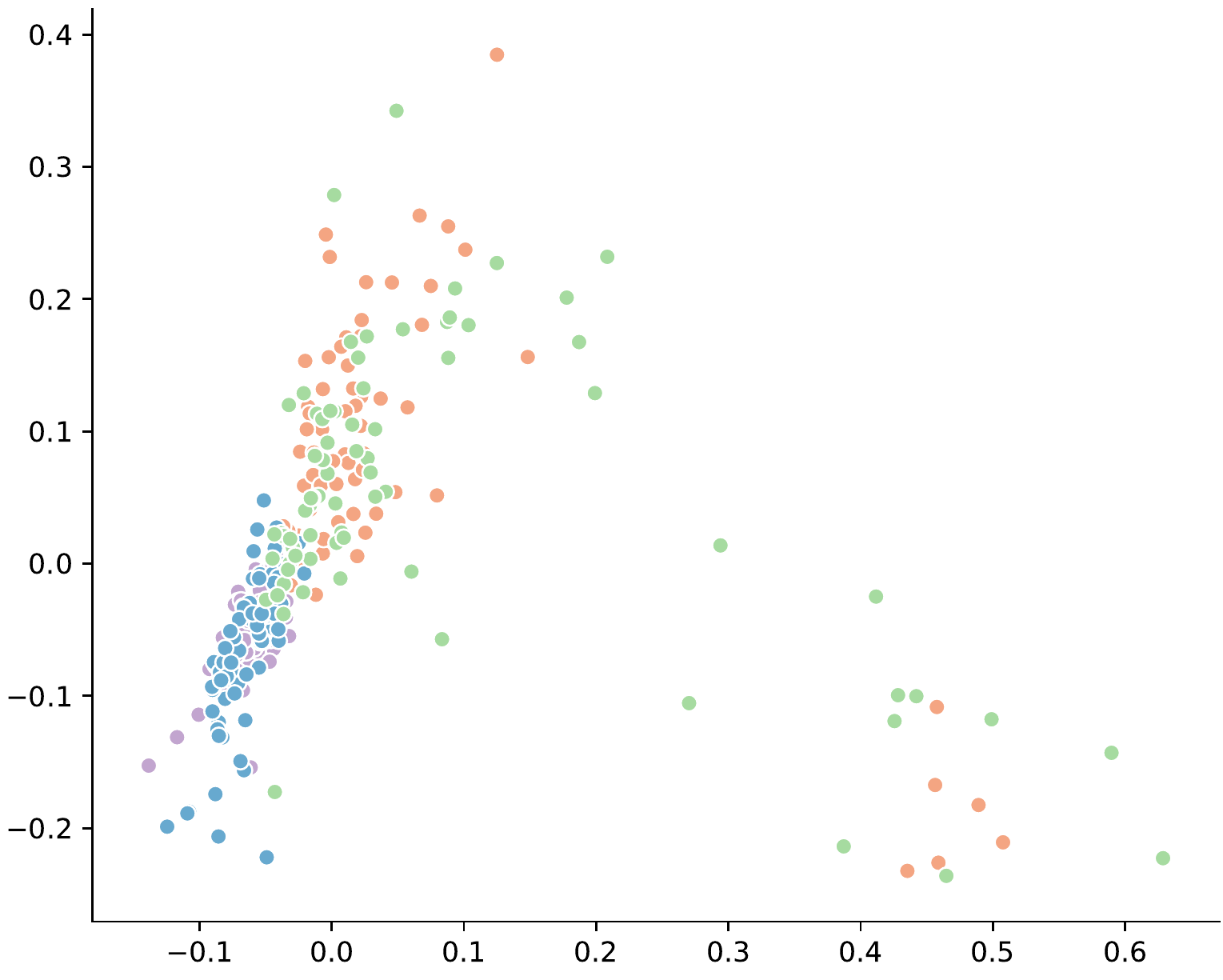}} \hfil
	\subfloat[Results for \dataset{Twenty newsgroup}]{\includegraphics[width=0.54\linewidth, keepaspectratio,page=3]{img/results_real.pdf}}\\
	\subfloat[\dataset{Eclipse bug report} dataset]{\includegraphics[width=0.4\linewidth, keepaspectratio,page=1]{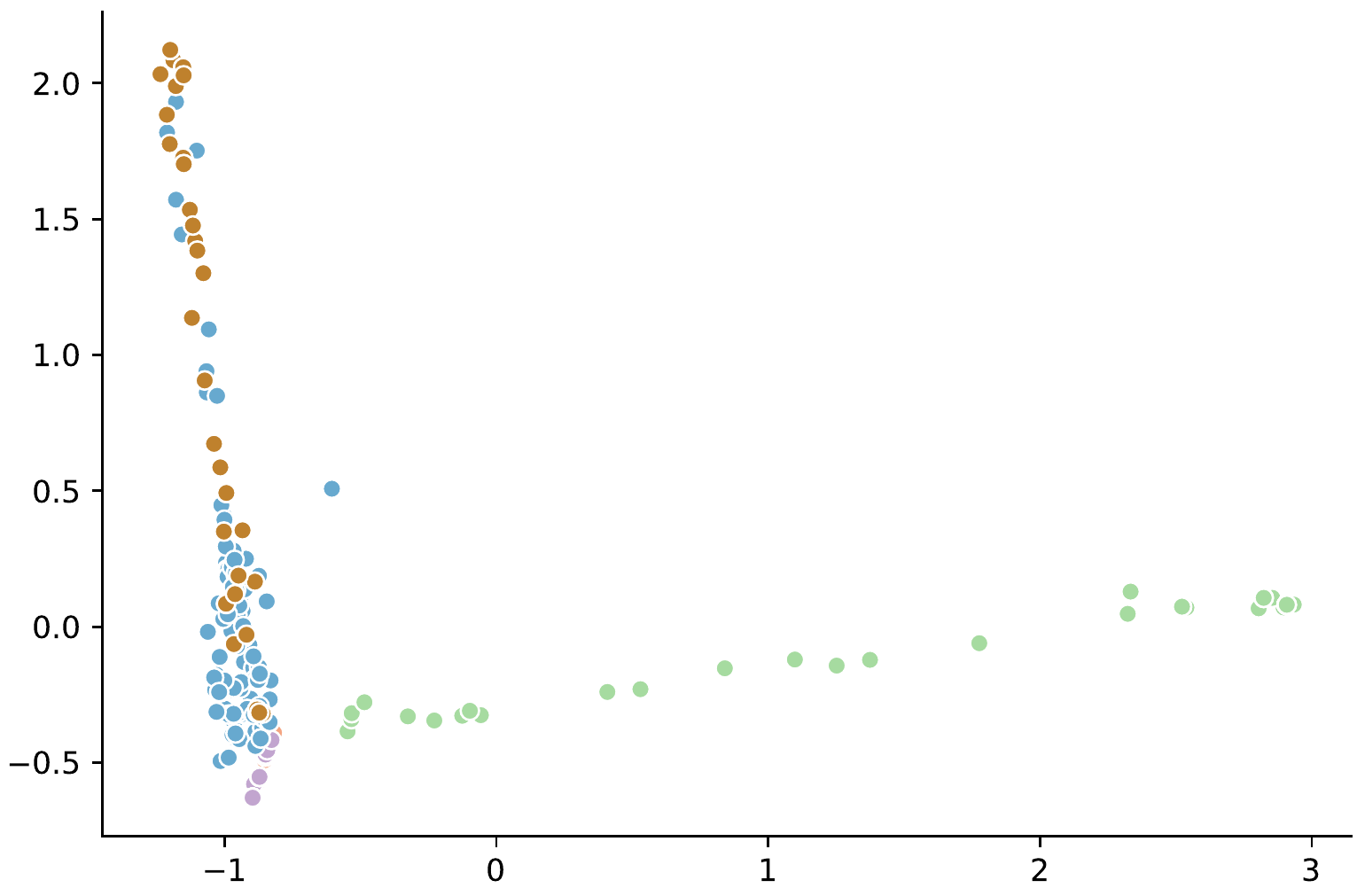}}\hfil
	\subfloat[Results for \dataset{Eclipse bug report} ]{\includegraphics[width=0.54\linewidth, keepaspectratio,page=4]{img/results_real.pdf}}
	\caption{ Ground-truth partition, fitness score and ARI for two text datasets.}
	\label{fig:results_text}
\end{figure}

The results are reported in Figure \ref{fig:results_text}. We observe that our fitness score was able to identify  the Cosine distance as an appropriate metric for clustering text documents. In addition, the Damerau–Levenshtein distance has given the highest fitness score for the \dataset{Eclipse bug report} instance. This demonstrates once again that the proposed fitness score is a valid tool 
to suggest a metric that adheres the most to the data.

\section{Maintaining geometrical properties of the dataset}
\label{sec:user_case2}

In this section, we demonstrate that dual information can be of great interest when experts are not entirely  sure of the pairwise constraints they provide.
In such circumstances, one may wish to perform distance metric learning more carefully, preserving as much as possible the characteristics of  the original data distribution while incorporating the domain knowledge embedded in the pairwise constraints. 

With that in mind, we propose to use 
dual information to establish an order in which pairwise constraints should be processed by metric learning algorithms. This sorting step aims to help these algorithms find the least impactful data transformations that can be applied,
that is, those transformations that require minimal modifications to the original data. 
Thus, our strategy aims not only to integrate the information contained in the pairwise constraints, but also to maintain a reliable representation of the data.

With MSSC as underlying clustering model, we designed a very simple distance metric learning algorithm that iteratively selects a violated constraint, and then performs data transformations necessary to satisfy it.
The method exploits the dual information in order to rank the pairwise constraints according to their estimated impact on the clustering objective function  \eqref{eq:clustering_obj}.

Algorithm \ref{alg:distance_learning} describes the steps of the proposed method. The algorithm begins by running an unsupervised clustering algorithm that generates an initial solution $X$ to the clustering problem \eqref{eq:clustering_obj}-\eqref{eq:clustering_end}, without constraints \eqref{eq:copcannotlink} and \eqref{eq:copmustlink} (Step \ref{algo:step3}). We then run the sub-gradient algorithm to solve \eqref{eq:dual_lagrange} and thus produce optimal dual values $\eta_{ij}^{c}, \lambda_{ij}^{c} $ and $ \gamma_{ij}^{c} $ (Step \ref{algo:step4}).  An {\emph impact score} $ \mathcal{I}_ {ij} $ is then calculated for
each pairwise constraint $(o_i,o_j) \in \mathcal{CL}\cup\mathcal{ML}$, this score being defined as the sum of the values of the dual variables which are associated with the constraint (Step \ref{algo:step5}).  Next, we
determine the pair $(o_i,o_j) \in \mathcal{CL}\cup\mathcal{ML}$ of maximum impact score $ \mathcal{I}_ {ij} $ whose associated constraint is violated by $ X $ (Step \ref{algo:step8}).
The reason behind this idea of maximizing the impact score is that the higher the value of a dual variable (i.e. the less negative it is), the lower will be the gain in the objective function if the associated constraint is relaxed. Thus, the data transformation required to satisfy such a constraint will presumably be small. We then run Algorithm~\ref{alg:satisfyng} that moves $o_i$ and/or $o_j$ in order to try to satisfy constraint $(o_i,o_j)$ (Step \ref{algo:step9}). Details on how this is done are given below.
This process is repeated until $X$ satisfies all constraints.

Note that the pairwise constraints can be conflicting and Algorithm \ref{alg:distance_learning} would therefore never stop. In addition, the algorithm can cycle. Indeed, suppose that $ k = 2 $ and that there are only two cannot-link constraints $ (o_1, o_2) $ and $ (o_1, o_3) $. If $ o_1 $ and $ o_2 $ belong to $ C_1 $ while $ o_3 $ belongs to $ C_2 $, the algorithm can possibly move $ o_1 $ towards $ C_2 $ to satisfy constraint $ (o_1, o_2) $. But $ (o_1, o_3) $ will then probably be violated, and the algorithm can possibly move $ o_1 $ back towards $ C_1 $, and this cycle can be repeated indefinitely. To avoid such situations, when moving a data point $ o_i \in C_j $ to a new position, we add the pair $ (i, j) $  in an initially empty list $ \mathcal{L} $ (Step \ref{algo:step10}), and we then forbid moving  $ o_i $ back to $ C_j  $. 
 More precisely, suppose constraint $(o_i,o_j)\in \mathcal{ML}$ is violated by $X$. At Step~\ref{algo:step6}, we define $M_{ij}$ as the set of indices $c$ of clusters such that the move of $o_i$ and $o_j$ towards $C_c$ is not forbidden (i.e., their move does not belong to $\mathcal{L}$). Similarly, for a violated constraint $(o_i,o_j)\in \mathcal{CL}$, we define $M_{ij}$ at Step \ref{algo:step7} as the set of indices $c$ of clusters such that $C_c$ does not contain $o_i$ and $o_j$ (i.e., $x_{i}^{c}{=}x_{j}^{c}{=}0$), and
the  move of at least one of the two data points towards cluster $C_c$ does not belong to $\mathcal{L}$. Let $\mathcal{M}$ be equal to the union of the sets $M_{ij}$ over all $(o_i, o_j) \in \mathcal{ML}\cup\mathcal{CL}$ (Step \ref{algo:new8}). The algorithm stops (Step \ref{algo:step11}) when $\mathcal{M}{=}\emptyset$, which means that $ X $ satisfies all the constraints, or  all moves which would satisfy a violated constraint are in $ \mathcal{L} $.
The
algorithm is finite since a data point $o_i$ is moved towards a cluster $C_c$ only if $(i,c){\notin} \mathcal{L}$, while $\mathcal{L}$ increases at each iteration.

\begin{algorithm}
	\caption{Using dual information for satifying all pairwise constraints}
	\label{alg:distance_learning}
	\begin{algorithmic}[1]
         \STATE Set $\mathcal{L}\leftarrow \emptyset$.\vspace{3pt}
		\REPEAT
	    \STATE Run an unsupervised MSSC algorithm for solving model \eqref{eq:clustering_obj}-\eqref{eq:clustering_end} without constraints \eqref{eq:copcannotlink} and \eqref{eq:copmustlink} and let $X= (x_{i}^{c})$ be the resulting solution.\vspace{2pt}\label{algo:step3}
	    \STATE Run the sub-gradient algorithm to solve \eqref{eq:dual_lagrange} and thus produce optimal dual values $\eta_{ij}^{c}, \lambda_{ij}^{c} $ and $ \gamma_{ij}^{c} $.\label{algo:step4}
			\STATE For all $(o_i,o_j) \in \mathcal{CL}\cup\mathcal{ML}$ do 
			$          {\mathcal{I}}_{ij} \leftarrow \begin{cases}
			\displaystyle\sum_{c=1}^k \eta_{ij}^{c}& \text{ if } (o_i, o_j) \in \mathcal{CL}  \\ 
			\displaystyle\sum_{c=1}^k (\lambda_{ij}^{c} + \lambda_{ij}^{\prime c}) & \text{ if } (o_i, o_j) \in \mathcal{ML}.\vspace{2pt}
		\end{cases}$\label{algo:step5}
    		\STATE For all $(o_i, o_j) \in \mathcal{ML}$ set $M_{ij}\leftarrow\{c\in\{1,\ldots,k\} \mid \{(i,c),(j,c)\}\cap \mathcal{L}=\emptyset\}$ if  $X$ violates $(o_i, o_j)$, and $M_{ij}\leftarrow \emptyset$ otherwise.\vspace{2pt}\label{algo:step6}
     		\STATE For all $(o_i, o_j) \in \mathcal{CL}$ set $M_{ij}\leftarrow\{c\in\{1,\ldots,k\} \mid x_{i}^{c}=x_{j}^{c}=0$ and $\{(i,c),(j,c)\}\nsubseteq \mathcal{L}\}$ if  $X$ violates $(o_i, o_j)$, and $M_{ij}\leftarrow \emptyset$ otherwise.\vspace{2pt}\label{algo:step7}
			\STATE Set $\mathcal{M}$ equal to the union of the sets $M_{ij}$ over all $(o_i, o_j) \in \mathcal{ML}\cup\mathcal{CL}$.\label{algo:new8}
			\IF{$\mathcal{M}\neq \emptyset$}
			\STATE Determine the pair $(i,j)$ that maximizes $\mathcal{I}_{ij}$ among those with $M_{ij}\neq \emptyset.$\label{algo:step8}
			\STATE Run Algorithm \ref{alg:satisfyng} to modify the coordinates of $o_i$ and/or $o_j$.\label{algo:step9}
 			\STATE Let $c_i$ and $c_j$  be the indices of the clusters such that $x_{i}^{c_i}=x_{j}^{c_j}=1$.\\
 			\textbf{if} the coordinates of $o_i$ have been changed at Step \ref{algo:step9} \textbf{then} add $(i,c_i)$ to $\mathcal{L}$.\\
 			\textbf{if} the coordinates of $o_j$ have been changed at Step \ref{algo:step9} \textbf{then} add $(j,c_j)$ to $\mathcal{L}$.\label{algo:step10}
 			\ENDIF
		\UNTIL{$\mathcal{M}= \emptyset$}.\label{algo:step11}
	\end{algorithmic}
\end{algorithm}

We now explain how Algorithm~\ref{alg:satisfyng} moves $o_i$ and/or $o_j$ in an attempt to satisfy a violated constraint $(o_i,o_j)\in\mathcal{ML}\cup\mathcal{CL}$.
Let us first define more precisely what we mean by moving a point $o_i\in C_c$ towards a new  cluster $C_{c^*}$. Such a move is obtained by modifying the coordinates of $o_i$ as follows. Let $\ell_1$ be the line containing $o_i$ and parallel to
$\overrightarrow{y_c,y_{c^*}}$, let $\ell_2$ be the perpendicular bisector of the segment tha connects $y_c$ with  $y_{c^*}$, and let $p$ be the point at the intersection of $\ell_1$ and $\ell_2$ : point $o_i$ is moved to the new location $p'$ such that 
$\overrightarrow{o_ip'}=\frac{101}{100}\overrightarrow{o_ip}$, which means that $o_i$ is then slightly closer to $y_{c^*}$ than to $y_{c}$. %{\blue Figure \ref{fig:scheme_transformation} illustrates such movement.}

%\begin{figure}[!htb]
 %   \centering
%    \includegraphics[width=0.7\linewidth, keepaspectratio]{img/move_towards_new_cluster.pdf}
%    \caption{Illustration associated to moving a point $o_i$ towards a new cluster $c^*$}
%    \label{fig:scheme_transformation}
%\end{figure}

As mentioned in Section \ref{sec:methodology}, the value $f(X)$ of a solution $X$ with MSSC as clustering criterion is
$$\sum_{i=1}^{n}\sum_{c=1}^{k} x_{i}^{c}\|o_i - y_c\|^{2}$$
where $y_c$ is the centroid of cluster $C_c$. As shown in ~\citep{HANSEN2001405}, if a new solution $X'$ is obtained from $X$ by moving a data point $o_i\in C_c$ to a new cluster $C_{c^*}$ with $c\neq c^*$ (i.e., $x_{i}^{c}=1$ and $x_{i}^{c^*}=0$ are replaced by $x_{i}^{c}=0$ and $x_{i}^{c^*}=1$), then the variation 
$\Delta(i,c, c^*)=f(X')-f(X)$ is equal to the following expression:
$$\Delta(i,c, c^*) = \frac{\sum_{i=1}^{n} x_i^{c^*}}{\sum_{i=1}^{n} x_i^{c^*}+1}\|o_i-y_{c^*}\|^2 - \frac{\sum_{i=1}^{n} x_i^{c}}{\sum_{i=1}^{n} x_i^{c}-1}\|o_i-y_{c}\|^2.
$$
Since we do not want to move $o_i$ towards $C_{c^*}$ if $(i,c^*)\in \mathcal{L}$, and $o_i$ does not move if $c=c^*$, we consider $\delta(o_i,c, c^*)$ defined as follows:
$$\delta(i,c, c^*)=\begin{cases}
\Delta(i,c, c^*)&\text{if }(i,c^*)\notin \mathcal{L}\text{ and }c\neq c^*\\
\infty&\text{if }(i,c^*)\in \mathcal{L}\text{ and }c\neq c^*\\
0&\text{if }c=c^*
\end{cases}$$

We are now ready to explain how Algorithm \ref{alg:satisfyng} works.
Let $(o_i,oj)$ be a violated constraints $\in\mathcal{CL}$ and assume that $o_i$ and $o_j$  currently  belong both to cluster $C_c$.
Algorithm~\ref{alg:satisfyng} determines the index $u\in\{i,j\}$  and the cluster index $c^*\neq c$ in $M_{ij}$ with minimum value $\delta(u,c, c^*)$, and then moves $o_u$ towards $C_{c^*}$, as explained above.

Similarly, let $(o_i,o_j)$ be a violated constraint $\in\mathcal{ML}$ and let $c_i$ and $c_j$ be the indices of the clusters that contain $o_i$ and $o_j$, respectively. Algorithm~\ref{alg:satisfyng} determines the cluster index $c^*\in M_{ij}$ with minimum value $\delta(i,c_i, c^*)+\delta(j,c_j, c^*)$. If $o_i\notin C_{c^*}$ then $o_i$ is moved towards $C_{c^*}$. Also, if $o_j\notin C_{c^*}$, then $o_j$ is moved towards $C_{c^*}$.

\begin{algorithm}
	\caption{Moving data points of violated constraints}
	\label{alg:satisfyng}
	\textbf{Input:} a violated constraint $ (o_i,o_j)\in\mathcal{ML}\cup\mathcal{CL}$.
	\begin{algorithmic}[1]
		\IF{$ (o_i,o_j) \in \mathcal{CL}$}
		    \STATE Determine $u{\in}\{i,j\}$  and the cluster index $c^*{\in} M_{ij}$ with minimum value $\delta(u,c, c^*)$.
		    \STATE Move $o_u$ towards $C_{c^*}$.
		\ELSE
		\STATE Let $c_i$ and $c_j$ be the indices of the clusters that contain $o_i$ and $o_j$, respectively.\\ 
		\STATE Determine the cluster index $c^*{\in}M_{ij}$ with minimum value $\delta(i,c_i, c^*){+}\delta(j,c_j, c^*)$. \\
        \STATE\textbf{If} $o_i\notin C_{c^*}$ \textbf{then} move $o_i$ towards $C_{c^*}$.\\
		\STATE\textbf{If} $o_j\notin C_{c^*}$ \textbf{then} move $o_j$  towards $C_{c^*}$.
\ENDIF
	\end{algorithmic}
\end{algorithm}

 We now demonstrate that algorithm \ref{alg:distance_learning} produces  high-quality clustering solutions while maintaining geometrical properties of the data as much as possible. For this purpose, we compare it to a baseline procedure that iteratively selects, at random, one violated constraint, and then performs the move defined in Algorithm \ref{alg:satisfyng}.

The algorithms are tested on three real-world datasets, namely the \dataset{Iris} dataset mentioned in Table \ref{tab:real_data}, the  \dataset{Seed} dataset that contains 210 samples with seven features, and the  \dataset{Optical Digits} dataset composed of 3823 handwritten digits images of 8×8 pixels.
These datasets were selected because (i) their ground-truth partitions are sufficiently different from the solutions obtained by optimizing the corresponding unsupervised MSSC model, and (ii) clustering them using the unsupervised MSSC model is still effective in terms of the obtained ARI indices, thereby reducing the number of necessary pairwise constraints to get close to the ground-truth partitions.

To get the solution $X$ in step 3 of Algorithm \ref{alg:distance_learning}, we executed 100 times the algorithm $k$-means, which is the most popular heuristic for solving unsupervised MSSC clustering problems, and we then set $X$ equal to the solution with minimum MSSC value.

For each datatset, we generated 20 constraint sets. All these sets are composed of pairwise constraints, randomly chosen among those that are violated by the initial solution $X$. For the \dataset{Iris}, instance, these sets contain 15 constraints, while  their number is 20 and 400 for the
\dataset{Seed} and \dataset{Optical} datasets, respectively. 

\begin{figure*}[!t]
	\centering
	\subfloat[ARI progression for \dataset{Iris}]{\includegraphics[width=0.5\linewidth, keepaspectratio,page=1]{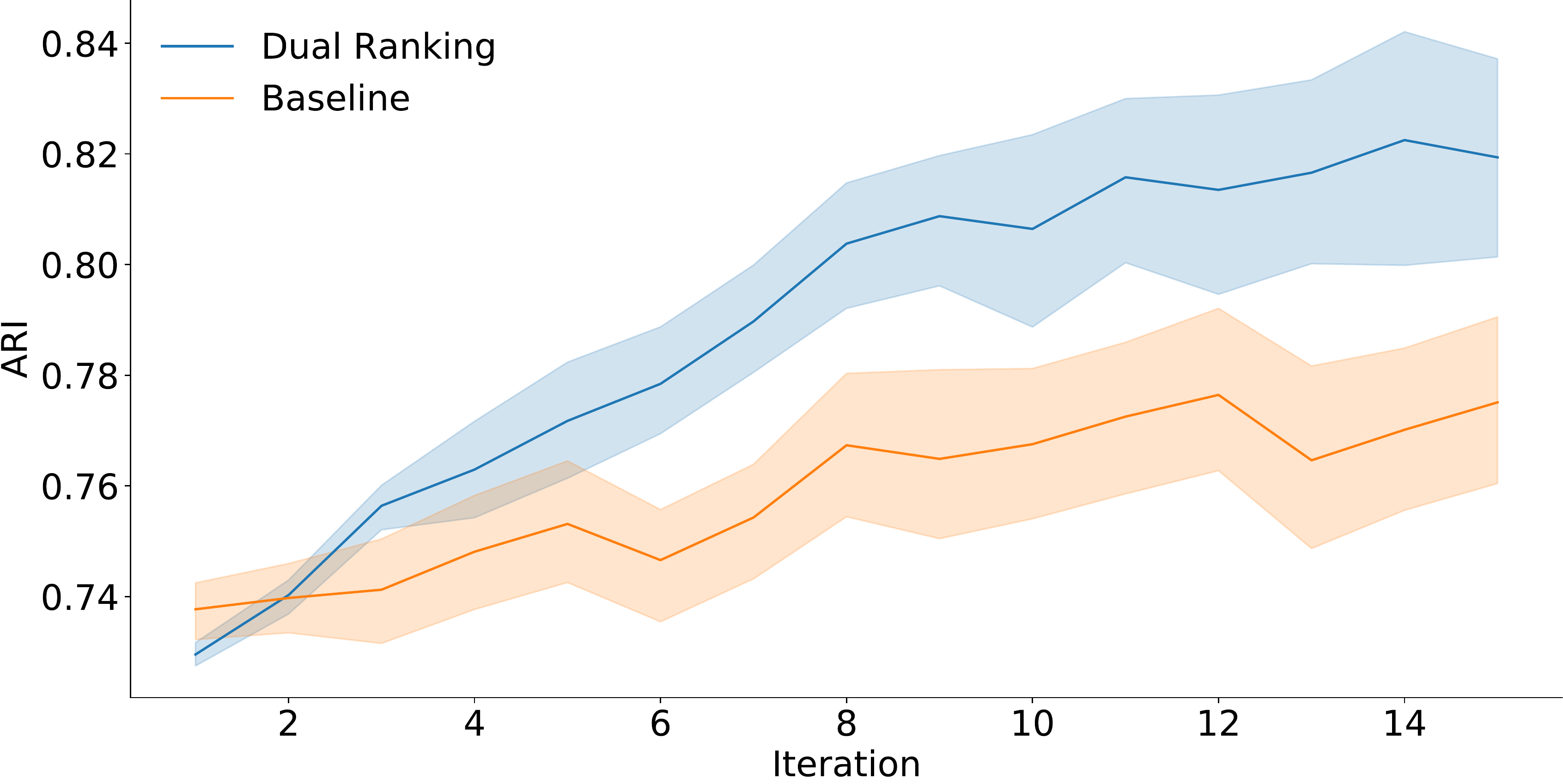}} \hfil
	\subfloat[Comulative traveled distance for \dataset{Iris}]{\includegraphics[width=0.5\linewidth, keepaspectratio,page=2]{img/distance_learning.pdf}} \\
	
	\subfloat[ARI progression for \dataset{Seed}]{\includegraphics[width=0.5\linewidth, keepaspectratio,page=3]{img/distance_learning.pdf}} \hfil
	\subfloat[Comulative traveled distance for \dataset{Seed}]{\includegraphics[width=0.5\linewidth, keepaspectratio,page=4]{img/distance_learning.pdf}} \\
	
	\subfloat[ARI progression for \dataset{Optical}]{\includegraphics[width=0.5\linewidth, keepaspectratio,page=5]{img/distance_learning.pdf}} \hfil
	\subfloat[Comulative traveled distance for \dataset{Optical}]{\includegraphics[width=0.5\linewidth, keepaspectratio,page=6]{img/distance_learning.pdf}}
	\caption{ARI progression and cumulative Euclidian traveled distance for Algorithm \ref{alg:distance_learning} and the baseline method applied to three real-world instances.}
	\label{fig:distance_learning}
\end{figure*}

At the end of each iteration, we  computed the ARI of the current solution $X$ as well as the cumulative Euclidean distance traveled by the transformed data points, these moves resulting from running Algorithm \ref{alg:satisfyng}. 
The results produced by Algorithm \ref{alg:distance_learning} and by the baseline method are reported in Figure \ref{fig:distance_learning}. The line and the band refer to the mean and standard deviation with respect to the 20 runs of each method.

Our first observation is that for each set of $m$  constraints ($ m = 15, 20 $ or $ 400 $), each execution of  Algorithm \ref{alg:distance_learning} only required $m$ iterations to satisfy them all, which means that each move resulting from a run of Algorithm \ref{alg:satisfyng} allowed to satisfy the violated constraint given in input. 

Next, for all datasets, we observe an upward progression of the ARI both for the proposed algorithm that exploits the dual information and for the baseline method. This behavior asserts the effectiveness of the designed distance metric learning method to leverage information from the provided pairwise constraints. More interestingly, we highlight that Algorithm \ref{alg:distance_learning} improves the clustering quality faster (except for \dataset{Seed}) and with less significant transformations of the original dataset when compared to the baseline algorithm that does not exploit dual information.

\section{Filtering  useful pairwise constraints}
\label{sec:user_case3}

Semi-supervised clustering techniques use constraints  to guide an algorithm towards better clustering solutions. However, some constraints can have a negative effect on the clustering task~\citep{Davidson2006}, and it is not necessarily easy to identify which constraints are useful or harmful to the clustering process  by basing this identification on a solution produced by an  algorithm.

We propose to use the information associated with the dual variables to rank the pairwise constraints according to the likelihood that they contribute favorably to the semi-supervised clustering process. We then only use the best ranked constraints, this selection aiming to reduce the number of constraints to consider without compromising the quality of the solution produced. As in the previous section, the pairwise constraints $(o_i,o_j)\in\mathcal{ML}\cup\mathcal{CL}$ are ordered according to their associated impact $\mathcal{I}_{ij}$.

A particular family of methods that can greatly benefit from this ranking approach are those designed to handle massive datasets, where the time required to process the data itself and the set of constraints is crucial for their use in the practice. We demonstrate the effectiveness of our approach by integrating it into the semi-supervised deep learning framework recently proposed by \citet{Zhang2021}. Their algorithm can be summarized as follows.

The algorithm first uses a deep neural network to determine 
an embedding from the data space to
a lower-dimensional feature space $Z$. The algorithm then computes soft and hard allocation values $q_{ic}$ and $p_{ic}$ in $[0,1]$ which, roughly speaking, can be interpreted as the probability
of assigning sample $o_i$ to cluster $C_c$. In the sequel, it defines a clustering loss function $\ell_C$ which is defined as 
$$\ell_C = \sum_{i=1}^{n}\sum_{c=1}^{k}p_{i}^{c}\log\frac{p_{i}^{c}}{q_{i}^{c}}, $$ \noindent as well as a loss $\ell_{\mathcal{ML}}$ for the must-link constraints, and a loss $\ell_{\mathcal{CL}}$ for the cannot-link constraints, where $\ell_{\mathcal{ML}}$ and $\ell_{\mathcal{CL}}$ are defined as
$$\ell_{\mathcal{ML}} = \sum_{(o_i,o_j)\in \mathcal{ML}} \log\sum_{c=1}^{k}q_{i}^{c}q_{j}^{c} \quad\text{ and }\quad \ell_{\mathcal{CL}} = \sum_{(o_i,o_j)\in \mathcal{CL}} \log(1-\sum_{c=1}^{k}q_{i}^{c}q_{j}^{c}). $$
The three loss functions $\ell_C$, $\ell_{ML}$ and $\ell_{CL}$ are then used to update the parameters of the neural network and to obtain a new mapping of the data points in $Z$. This process is repeated at most $m$ times, where $m$ is a parameter, but may stop earlier if the ratio of changed cluster assignments between two consecutive iterations is smaller than $0.001$, each data point $o_i$ being assigned to the cluster $c$ that maximizes $q_i^c$.
For more details, the reader is referred to \cite{Zhang2021}.

Instead of defining $ \ell_{ML} $ and $\ell_{CL} $ by considering all the available pairwise constraints, we propose to use only a subset. More precisely, we run the sub-gradient algorithm to determine optimal dual values $\eta_{ij}^{c}, \lambda_{ij}^{c} $ and $ \gamma_{ij}^{c} $ for the clustering problem based on the embedding of the points in $Z$. As in the previous section, these dual values are used to define the impact score $ \mathcal{I}_{ij} $ of each constraint $(o_i,o_j)\in\mathcal{ML}\cup\mathcal{CL}$, where $ \mathcal{I}_{ij} $ is defined as the sum of the values of the dual variables associated with that constraint. 
Let $\Omega$ be the subset of pairwise constraints having a strictly negative impact : we eliminate the $\lfloor \alpha\rfloor|\Omega|$ constraints of $\Omega$ which have the smallest impact score, where $\alpha\in[0,1]$ is a parameter, and we denote by $\Omega^{\alpha}$ the set of remaining constraints. By defining $\mathcal{ML}^{\alpha}=\mathcal{ML}\cap\Omega^{\alpha}$ and $\mathcal{CL}^{\alpha}=\mathcal{CL}\cap \Omega^{\alpha}$, we obtain the following new loss functions $\ell_{\mathcal{ML}}^{\alpha}$ and  $\ell_{\mathcal{CL}}^{\alpha}$ given by :
$$\ell_{\mathcal{ML}}^{\alpha} = \sum_{(o_i,o_j)\in \mathcal{ML}^{\alpha}} \log\sum_{c=1}^{k}q_{i}^{c}q_{j}^{c} \quad\text{ and }\quad \ell_{\mathcal{CL}}^{\alpha} = \sum_{(o_i,o_j)\in \mathcal{CL}^{\alpha}} \log(1-\sum_{c=1}^{k}q_{i}^{c}q_{j}^{c}). $$

The neural network parameters are updated by using $\ell_{ML}^{\alpha}$ and  $\ell_{CL}^{\alpha}$ instead of $\ell_{ML}$ and  $\ell_{CL}$. Algorithm \ref{alg:dec} describes the main steps of this integration of our dual approach in the algorithm of \cite{Zhang2021}.

\begin{algorithm}[!htb]
	\centering
	\caption{Integration of the dual approach in the deep learning framework of \cite{Zhang2021}}
	\label{alg:dec}
	\begin{algorithmic}[1]
		 \STATE Train the neural network to obtain an embedding of the data points in a lower dimensional feature space $ Z  $. Initialize the iteration counter $iter$ to zero. \vspace{2pt}
		\REPEAT\vspace{2pt}
		\STATE Compute the soft and hard allocation values $ q^{c}_{i}$ and $p_{i}^{c}$ as well as the loss $\ell_C$.\vspace{2pt}
		\STATE Run the sub-gradient algorithm to produce optimal dual values $\eta_{ij}^{c}, \lambda_{ij}^{c} $ and $ \gamma_{ij}^{c}$ for the clustering problem based on the current embedding of the data points in  $Z$.\vspace{2pt}
		\STATE For all $(o_i,o_j) \in \mathcal{CL}\cup\mathcal{ML}$ do 
			$          {\mathcal{I}}_{ij} \leftarrow \begin{cases}
			\displaystyle\sum_{c=1}^k \eta_{ij}^{c}& \text{ if } (o_i, o_j) \in \mathcal{CL}  \\ 
			\displaystyle\sum_{c=1}^k (\lambda_{ij}^{c} + \lambda_{ij}^{\prime c}) & \text{ if } (o_i, o_j) \in \mathcal{ML}.\vspace{2pt}
		\end{cases}$.\vspace{2pt}
		\STATE Set $\Omega {\leftarrow} \{ (o_i,o_j){\in} \mathcal{ML}\cup \mathcal{CL} \; | \; \mathcal{I}_{ij}{<} 0\}$, sort these constraints in increasing order of their impact score $\mathcal{I}_{ij}$, and remove the $\lfloor \alpha\rfloor|\Omega|$ top-ranked constraints of $\Omega$ to obtain $\Omega^{\alpha}$.\vspace{2pt}
		\STATE Set $\mathcal{ML}^{\alpha}\leftarrow\mathcal{ML}\cap\Omega^{\alpha}$, $\mathcal{CL}^{\alpha}\leftarrow\mathcal{CL}\cap\Omega^{\alpha}$, and compute the losses $\ell_{ML}^{\alpha}$and $\ell_{CL}^{\alpha}$.\vspace{2pt}
        \STATE Set $iter\leftarrow iter{+}1$, update the neural network parameters by using $\ell_C$, $\ell_{CL}^{\alpha}$ and $\ell_{ML}^{\alpha}$, and get a new embedding of the data points in $Z$.\vspace{2pt}
		\UNTIL $iter{=}m$ or the ratio of changed cluster assignments between two consecutive iterations is below $10^{-3}$, each $o_i$ being assigned to the cluster $c$ that maximizes $q_i^c$.
	\end{algorithmic}

\end{algorithm}

 Algorithm \ref{alg:dec} is compared with the original algorithm of \citet{Zhang2021} on the same three datasets they used for their experiment: \dataset{MNIST} which contains 60,000 handwritten digits of $28{\times} 28$ pixel size which have to be grouped into $ k = 10 $ classes, \dataset{Fashion-MNIST} which contains 60,000 grayscale $28{\times} 28$
images and $ k = 10 $ classes, and \dataset{Reuters-10K} which contains 10,000 document texts with 2,000 features and $k= 10$ classes. For each dataset, we generated 20 sets of  pairwise constraints randomly chosen from the ground-truth partition. The number of constraints in each set is equal to $10\%$ of the size of the dataset, which amounts to 6,000 for \dataset{MNIST} and \dataset{Fashion-MNIST}, and 1,000 for \dataset{Reuters-10K}. We ran Algorithm \ref{alg:dec} for at most $m=500$ iterations, with $\alpha=0.95, 0,9, 0.8, 0.7, 0.6$ and $0.0$. Note that running our algorithm with $\alpha{=}0.0$ is not totally equivalent to applying the original algorithm of \citet{Zhang2021} since we do not consider pairwise constraints $(o_i,o_j)$ with impact score $\mathcal{I}_{ij}=0$.

We indicate in Table \ref{tab:results_filtering} the average ARI (mean $\pm$ standard deviation) over the 20 constraint sets  for each value of $\alpha$ and for the three  datasets. CPU times (in seconds) and the number of iterations required to reach convergence are shown in Table \ref{tab:epochs_result_filtering} .
 We observe that our strategy that uses the dual information for filtering the constraints improves the unfiltered version of \citet{Zhang2021} for all three datasets. The best ARI is obtained by removing 70 to $80\%$ of the constraints with the lowest impact score. Note also that in addition to improving the ARI,  our filtering approach does not significantly modify the CPU times and the number of iterations necessary to reach convergence. More precisely, a reduction is obtained, both in terms of time and number of iterations, for the datasets \dataset{MNIST} and \dataset{Fashion-MNIST} when using our filtering algorithm with $ \alpha=70\% $. However, such gains are not observed for the smallest dataset \dataset{Reuters-10k}, where the unfiltered algorithm converges faster than Algorithm \ref{alg:dec}.

\begin{table}[!t]
	\caption{ Mean ARI and standard deviations for Algorithm \ref{alg:dec} and for the unfiltered original algorithm of \citet{Zhang2021}.}
	\centering
	\label{tab:results_filtering}
\begin{tabular}{@{}rccc@{}}
\toprule
         & \multicolumn{3}{c}{Datasets} \\ \cmidrule{2-4}
        \multicolumn{1}{r}{$\alpha$} & \multicolumn{1}{r}{MNIST} & \multicolumn{1}{r}{Fashion-MNIST} & \multicolumn{1}{r}{Reuters-10K} \\ \midrule
        0.95         & $0.9370 \pm 0.0136 $ & $ 0.4175\pm 0.0244$ &  $0.5992 \pm 0.0140$ \\
        0.90        & $0.9389 \pm  0.0083$ & $ 0.4115\pm 0.0193$ & $0.6006 \pm 0.0157$ \\
        0.80        & $ 0.9338\pm 0.0132$ & $\mathbf{0.4237 \pm 0.0250}$ & $\mathbf{0.6009 \pm 0.0153}   $ \\
        0.70       & $ \mathbf{0.9427 \pm 0.0077}$ & $ 0.4132\pm 0.0127$ & $0.5949 \pm 0.0154$ \\
        0.60       & $ 0.9343 \pm 0.0244$ & $ 0.4164\pm 0.0101$ & $0.5843 \pm 0.0153$ \\
        0.00       & $ 0.9343\pm 0.0161$ & $0.4012 \pm 0.0167$ & $0.5830 \pm 0.0120$ \\
        Unfiltered  & $ 0.9314 \pm 0.0040$ & $ 0.3916 \pm 0.0163$ & $0.5995 \pm 0.0108$  \\ \bottomrule
    \end{tabular}
\end{table}

\begin{table}[!t]
	\caption{CPU times (in seconds) and number of iterations to reach convergence for Algorithm \ref{alg:dec} and for the unfiltered original algorithm of \citet{Zhang2021}.}
	\centering
	\label{tab:epochs_result_filtering}
\setlength{\tabcolsep}{3pt}
	\resizebox{\textwidth}{!}{
	\begin{tabular}{@{}rcccccc@{}}
        \toprule
         & \multicolumn{2}{c}{MNIST} & \multicolumn{2}{c}{Fashion-MNIST} & \multicolumn{2}{c}{Reuters-10K}\\
         \cmidrule(lr){2-3} \cmidrule(lr){4-5} \cmidrule(lr){6-7}
        $\alpha$ & \multicolumn{1}{c}{CPU times} & \multicolumn{1}{c}{iterations} & \multicolumn{1}{c}{CPU times} & \multicolumn{1}{c}{iterations} & \multicolumn{1}{c}{CPU times} & \multicolumn{1}{c}{iterations}\\ \midrule
0.95	&$	171.7	\pm	22.6	$&$	126.5	\pm	16.4	$&$	646.4	\pm	19.5	$&$	499.3	\pm	2.6	$&$	33.4	\pm	5.0	$&$	67.1	\pm	18.2	$ \\
0.90	&$	152.5	\pm	17.9	$&$	127.7	\pm	15.4	$&$	658.7	\pm	05.1	$&$	497.1	\pm	8.8	$&$	34.2	\pm	3.8	$&$	61.8	\pm	14.8	$ \\
0.80	&$	161.0	\pm	33.1	$&$	134.9	\pm	29.0	$&$	673.4	\pm	23.9	$&$	465.3	\pm	37.1	$&$	30.0	\pm	1.9	$&$	50.1	\pm	08.1	$ \\
0.70	&$	\mathbf{142.8	\pm	23.8}	$&$	\mathbf{120.8	\pm	20.5}	$&$	\mathbf{587.5	\pm	48.7}	$&$	\mathbf{463.2	\pm	53.5}	$&$	29.3	\pm	2.0	$&$	44.8	\pm	06.3	$ \\

0.60	&$	164.3	\pm	33.4	$&$	128.2 \pm 26.2	$&$	591.0 \pm 18.1	$&$	497.9 \pm 09.1	$&$	28.8	\pm	1.4	$&$	34.4 \pm 06.2	$ \\

0.00	&$	183.6	\pm	22.9	$&$	137.4	\pm	20.2	$&$	623.8	\pm	40.3	$&$	488.3	\pm	26.5	$&$	26.6	\pm	1.2	$&$	28.6	\pm	06.3	$ \\
Unfiltered	&$	187.2	\pm	27.5	$&$	149.7	\pm	14.8	$&$	775.3	\pm	27.4	$&$	500.0	\pm	00.0	$&$	\mathbf{13.2	\pm	0.4}	$&$	\mathbf{10.7	\pm	01.9}	$\\
\bottomrule
\end{tabular}
}
\end{table}

\section{Conclusion}

Our goal was to explore how dual information can improve distance metric learning algorithms for clustering problems with must-link and cannot-link constraints. We have shown that the optimal dual values associated to these constraints
in a clustering problem are an effective tool to estimate the impact of violating a constraint. These dual values are obtained by running a sub-gradient algorithm on a Lagrangian relaxation of the original clustering problem, where pairwise constraints are replaced by penalty terms in the objective function. We have illustrated this benefit for the following three tasks.

\begin{enumerate}
    \item  \textbf{Identification of an appropriate dissimilarity measure}. We have defined a fitness score based on optimal dual values for recommending to experts a dissimilarity measure that best agrees with their beliefs and expectations.
    
    \item \textbf{Preserving geometrical properties of the dataset}.
    Distance metric learning algorithms aim to move closer pairs of data points involved in must-link constraints, and to move pairs of points involved in cannot-link constraints away from each other.
   These transformations are often applied without worrying too much about their magnitude. We have shown how the use of dual information makes it possible to determine transformations with little impact on the original space. Our methodology therefore offers a tool to data analysts to allow them to integrate their knowledge of the domain while controlling the way in which the data is modified.
    
    \item \textbf{Filtering useful pairwise constraints}. 
    While it is known that some pairwise constraints can have a negative effect on the clustering task, the identification of these harmful constraints is not an easy task.
 We have defined an impact score, based on optimal dual values, which helps to filter the constraints that appear to be the most useful.
 We have demonstrated the benefit of this filter by integrating it into the recently proposed deep learning  framework of \citet{Zhang2021}.
\end{enumerate}

\begin{acknowledgements}
	This research was enabled in part by support provided by Calcul Qu\'ebec (\url{https://www.calculquebec.ca}) and Compute Canada (\url{www.computecanada.ca}).
\end{acknowledgements}


%\bibliographystyle{elsarticle-num-names}
\bibliographystyle{spbasic}      % basic style, author-year citations
\bibliography{mybibfile}   % name your BibTeX data base

\begin{thebibliography}{35}
\providecommand{\natexlab}[1]{#1}
\providecommand{\url}[1]{{#1}}
\providecommand{\urlprefix}{URL }
\expandafter\ifx\csname urlstyle\endcsname\relax
  \providecommand{\doi}[1]{DOI~\discretionary{}{}{}#1}\else
  \providecommand{\doi}{DOI~\discretionary{}{}{}\begingroup
  \urlstyle{rm}\Url}\fi
\providecommand{\eprint}[2][]{\url{#2}}

\bibitem[{Aggarwal(2015)}]{Aggarwal2015}
Aggarwal CC (2015) {Similarity and distances}, Springer International
  Publishing, Cham, pp 63--91. \doi{10.1007/978-3-319-14142-8_3}

\bibitem[{Aloise et~al.(2012)Aloise, Hansen, and Liberti}]{Aloise2012}
Aloise D, Hansen P, Liberti L (2012) {An improved column generation algorithm
  for minimum sum-of-squares clustering}. Mathematical Programming
  131(1):195--220, \doi{10.1007/s10107-010-0349-7}

\bibitem[{Baghshah and Shouraki(2009)}]{Baghshah2008}
Baghshah MS, Shouraki SB (2009) {Semi-supervised metric learning using pairwise
  constraints}. In: IJCAI International Joint Conference on Artificial
  Intelligence, pp 1217--1222

\bibitem[{Bar-Hillel et~al.(2005)Bar-Hillel, Hertz, Shental, and
  Weinshall}]{Aharon2005}
Bar-Hillel A, Hertz T, Shental N, Weinshall D (2005) Learning a {M}ahalanobis
  metric from equivalence constraints. Journal of Machine Learning Research
  6(32):937--965

\bibitem[{Bellet et~al.(2015)Bellet, Habrard, and Sebban}]{bellet2015metric}
Bellet A, Habrard A, Sebban M (2015) Metric learning. Synthesis Lectures on
  Artificial Intelligence and Machine Learning 9(1):1--151

\bibitem[{Bertsimas and Tsitsiklis(1997)}]{Bertsimas1997}
Bertsimas D, Tsitsiklis J (1997) Introduction to linear optimization, 1st edn.
  Athena Scientific

\bibitem[{Bilenko et~al.(2004)Bilenko, Basu, and Mooney}]{Bilenko2004}
Bilenko M, Basu S, Mooney RJ (2004) {Integrating constraints and metric
  learning in semi-supervised clustering}. Proceedings, Twenty-First
  International Conference on Machine Learning, ICML 2004 pp 81--88,
  \doi{10.1145/1015330.1015360}

\bibitem[{Breiman(1996)}]{Breiman1996}
Breiman L (1996) Bagging predictors. Machine Learning 24(2):123--140,
  \doi{10.1007/BF00058655}

\bibitem[{Chang and Yeung(2006)}]{Chang2006}
Chang H, Yeung DY (2006) {Locally linear metric adaptation with application to
  semi-supervised clustering and image retrieval}. Pattern Recognition
  39(7):1253--1264, \doi{10.1016/j.patcog.2005.12.012}

\bibitem[{Davidson et~al.(2006)Davidson, Wagstaff, and Basu}]{Davidson2006}
Davidson I, Wagstaff KL, Basu S (2006) Measuring constraint-set utility for
  partitional clustering algorithms. In: F{\"u}rnkranz J, Scheffer T,
  Spiliopoulou M (eds) Knowledge Discovery in Databases: PKDD 2006, Springer
  Berlin Heidelberg, Berlin, Heidelberg, pp 115--126

\bibitem[{Dua and Graff(2017)}]{Dua2019}
Dua D, Graff C (2017) {UCI} machine learning repository.
  \urlprefix\url{http://archive.ics.uci.edu/ml}

\bibitem[{Fisher(2004)}]{Fisher2004}
Fisher ML (2004) The {L}agrangian relaxation method for solving integer
  programming problems. Management Science 50(12\_supplement):1861--1871,
  \doi{10.1287/mnsc.1040.0263}

\bibitem[{Fogel et~al.(2019)Fogel, Averbuch-Elor, Cohen-Or, and
  Goldberger}]{Fogel2019}
Fogel S, Averbuch-Elor H, Cohen-Or D, Goldberger J (2019) Clustering-driven
  deep embedding with pairwise constraints. IEEE Computer Graphics and
  Applications 39(4):16--27, \doi{10.1109/MCG.2018.2881524}

\bibitem[{Hansen and Jaumard(1997)}]{Hansen1997}
Hansen P, Jaumard B (1997) {Cluster analysis and mathematical programming}.
  Mathematical Programming 79(1):191--215, \doi{10.1007/BF02614317}

\bibitem[{Hansen and Mladenović(2001)}]{HANSEN2001405}
Hansen P, Mladenović N (2001) J-means: a new local search heuristic for
  minimum sum of squares clustering. Pattern Recognition 34(2):405--413,
  \doi{https://doi.org/10.1016/S0031-3203(99)00216-2}

\bibitem[{Hsu and Kira(2015)}]{hsu2015neural}
Hsu YC, Kira Z (2015) {Neural network-based clustering using pairwise
  constraints}. Arxiv.org/abs/1511.06321, \eprint{1511.06321}

\bibitem[{Hubert and Arabie(1985)}]{Hubert1985}
Hubert L, Arabie P (1985) Comparing partitions. Journal of Classification
  2(1):193--218, \doi{10.1007/BF01908075}

\bibitem[{Kalintha et~al.(2017)Kalintha, Ono, Numao, and Fukui}]{Kalintha2017}
Kalintha W, Ono S, Numao M, Fukui KI (2017) {Kernelized evolutionary distance
  metric learning for semi-supervised clustering}. In: 31st AAAI Conference on
  Artificial Intelligence, AAAI 2017, vol~3, pp 4945--4946,
  \doi{10.3233/ida-184283}

\bibitem[{Kaufman and Rousseeuw(2009)}]{kaufman2009finding}
Kaufman L, Rousseeuw PJ (2009) Finding groups in data: an introduction to
  cluster analysis, vol 344. John Wiley \& Sons

\bibitem[{Kullback and Leibler(1951)}]{kullback1951}
Kullback S, Leibler RA (1951) On information and sufficiency. Ann Math Statist
  22(1):79--86, \doi{10.1214/aoms/1177729694}

\bibitem[{{Le Capitaine}(2018)}]{Lecapitaine2018}
{Le Capitaine} H (2018) Constraint selection in metric learning.
  Knowledge-Based Systems 146:91--103,
  \doi{https://doi.org/10.1016/j.knosys.2018.01.026}

\bibitem[{Lei et~al.(2009)Lei, Jin, Hoit, Zhu, and Yu}]{Lei2009}
Lei W, Jin R, Hoit SC, Zhu J, Yu N (2009) Learning {B}regman distance functions
  and its application for semi-supervised clustering. In: Advances in Neural
  Information Processing Systems 22 - Proceedings of the 2009 Conference, pp
  2089--2097

\bibitem[{Mahalanobis(1936)}]{maha}
Mahalanobis PC (1936) On the generalized distance in statistics. Proceedings of
  the National Institute of Sciences (Calcutta) 2:49--55

\bibitem[{Manning and Schutze(1999)}]{manning1999foundations}
Manning C, Schutze H (1999) Foundations of statistical natural language
  processing. MIT press

\bibitem[{Nguyen and Baets(2019)}]{Nguyen2019}
Nguyen B, Baets BD (2019) Kernel-based distance metric learning for supervised
  $k$ -means clustering. IEEE Transactions on Neural Networks and Learning
  Systems 30(10):3084--3095, \doi{10.1109/TNNLS.2018.2890021}

\bibitem[{Pichery(2014)}]{PICHERY2014236}
Pichery C (2014) Sensitivity analysis. In: Wexler P (ed) Encyclopedia of
  Toxicology (Third Edition), third edition edn, Academic Press, Oxford, pp 236
  -- 237, \doi{https://doi.org/10.1016/B978-0-12-386454-3.00431-0}

\bibitem[{Randel et~al.(2021)Randel, Aloise, Blanchard, and
  Hertz}]{randel2020a}
Randel RA, Aloise D, Blanchard SJ, Hertz A (2021) A {L}agrangian-based score
  for assessing the quality of pairwise constraints in semi-supervised
  clustering. Data Mining and Knowledge Discovery

\bibitem[{Rao(2013)}]{Rao2013}
Rao RV (2013) A novel weighted Euclidean distance-based approach, Springer
  London, London, pp 159--191. \doi{10.1007/978-1-4471-4375-8_5}

\bibitem[{Rodrigues et~al.(2020)Rodrigues, Aloise, and Fernandes}]{muller2020a}
Rodrigues I, Aloise D, Fernandes ER (2020) An effective sequence alignment
  method for duplicate crash report detection. In: 4th International Workshop
  on Machine Learning Techniques for Software Quality Evolution (MaLTeSQuE2020)

\bibitem[{Wu et~al.(2012)Wu, Hoi, Jin, Zhu, and Yu}]{Wu2012}
Wu L, Hoi SC, Jin R, Zhu J, Yu N (2012) Learning {B}regman distance functions
  for semi-supervised clustering. IEEE Transactions on Knowledge and Data
  Engineering 24(3):478--491, \doi{10.1109/TKDE.2010.215}

\bibitem[{Xiang et~al.(2008)Xiang, Nie, and Zhang}]{Xiang2008}
Xiang S, Nie F, Zhang C (2008) Learning a {M}ahalanobis distance metric for
  data clustering and classification. Pattern Recognition 41(12):3600--3612,
  \doi{10.1016/j.patcog.2008.05.018}

\bibitem[{Xie et~al.(2016)Xie, Girshick, and Farhadi}]{xie2016unsupervised}
Xie J, Girshick R, Farhadi A (2016) Unsupervised deep embedding for clustering
  analysis. \eprint{1511.06335}

\bibitem[{Xing et~al.(2003)Xing, Jordan, Russell, and Ng}]{Xing2003}
Xing EP, Jordan MI, Russell SJ, Ng AY (2003) Distance metric learning with
  application to clustering with side-information. Advances in Neural
  Information Processing Systems 15 pp 521--528

\bibitem[{Zhang et~al.(2008)Zhang, Chen, and Zhou}]{Zhang2008}
Zhang D, Chen S, Zhou ZH (2008) Constraint score: a new filter method for
  feature selection with pairwise constraints. Pattern Recognition
  41(5):1440--1451, \doi{https://doi.org/10.1016/j.patcog.2007.10.009}

\bibitem[{Zhang et~al.(2021)Zhang, Zhan, Basu, and Davidson}]{Zhang2021}
Zhang H, Zhan T, Basu S, Davidson I (2021) A framework for deep constrained
  clustering. Data Mining and Knowledge Discovery 35(2):593--620,
  \doi{10.1007/s10618-020-00734-4}

\end{thebibliography}
\end{document}